\documentclass[letterpaper]{article} 
\usepackage{aaai25}  
\usepackage{times}  
\usepackage{helvet}  
\usepackage{courier}  
\usepackage[hyphens]{url}  
\usepackage{graphicx} 
\usepackage{natbib}  
\usepackage{caption} 
\usepackage{amsmath,amssymb}
\usepackage[utf8]{inputenc} 
\usepackage[T1]{fontenc}    
\usepackage{url}            
\usepackage{booktabs}       
\usepackage{amsfonts}       
\usepackage{nicefrac}       
\usepackage{microtype}      
\usepackage{xcolor}         
\usepackage{subcaption,moreverb,marvosym}
\usepackage{algpseudocode}
\usepackage{float}
\newcommand{\added}[1]{\textcolor{black}{#1}}
\title{Transformer Layers as Painters}

\author{
    Qi Sun\equalcontrib\textsuperscript{\rm 2,\rm 3}, 
    Marc Pickett\equalcontrib\thanks{Corresponding author.}\textsuperscript{\rm 1}, 
    Aakash Kumar Nain\textsuperscript{\rm 1}, 
    Llion Jones\textsuperscript{\rm 2}
}

\affiliations{
    \textsuperscript{1}Emergence AI\\
    \textsuperscript{2}Sakana AI, Japan\\
    \textsuperscript{3}Institute of Science Tokyo, Japan\\
    \{mpickett,anain\}@emergence.ai,
    \{qisun, llion\}@sakana.ai
}

\begin{document}\maketitle
\begin{abstract}
  Despite their nearly universal adoption for large language models, the internal workings of transformers are not well understood.
  We aim to better understand the impact of removing or reorganizing information throughout the layers of a pretrained transformer.
  Such an understanding could both yield better usage of existing models as well as to make architectural improvements to produce new variants.
  We present a series of empirical studies on frozen models that show that the lower and final layers of pretrained transformers differ from middle layers,
  but that middle layers have a surprising amount of uniformity.  We further show that some classes of problems have robustness to
  skipping layers, running the layers in an order different from how they were trained, or running the layers in parallel.
  Our observations suggest that even frozen pretrained models may gracefully trade accuracy for latency by skipping layers or running layers in
  parallel.
\end{abstract}

\begin{links}
\link{Code}{https://github.com/floatingbigcat/transformer_layers_as_painters}
\end{links}

\vspace{-2mm}
\section{Introduction}

The scale of transformer-based Large Language Models (LLMs), in the billions of parameters, makes it difficult to directly understand the
models' behaviour after training.  At the same time, each layer of a pretrained transformer has an identical architecture as the other layers,
with the only difference being a layer's position in the hierarchy, and the values of the layer's parameters \cite{Vaswani+2017}.

We find it helpful to think of the middle layers of a transformer by making an analogy to an assembly line of painters.
The canvas (input) is passed along a series of painters.  Some painters specialize in birds, while others are better at painting wheels.  Each
painter receives the canvas from the painter below her, then she decides whether to add a few strokes to the painting or just pass it along to
the painter above her (using the residual connections).
In this analogy, each painter uses the same ``vocabulary'' for understanding paintings, so that a painter may receive the painting from a
painter earlier in the assembly line without catastrophe.  The painters may also be reordered without complete catastrophe (even if parts of the
background get painted \emph{after} foreground objects, occluding them), and the painters may even all add their strokes at the same time (in
parallel).

This analogy isn't meant to be a rigorous theory, but rather a tool for thinking about a transformer's layers.
Inspired by this analogy, we test how well some hypotheses hold.
In this paper we perform experiments that help address the following questions:
\begin{enumerate} 
\item Do layers use the same representation space? (\S \ref{subsection:samelang})
\item Are all the layers necessary? (\S \ref{subsection:necessary})
\item Are middle layers all doing the same function? (\S \ref{subsection:samething})
\item Does the layer order matter? (\S \ref{subsection:order})
\item Can we run the layers in parallel? (\S \ref{subsection:parallel})
\item Does order matter for some tasks more than others? (\S \ref{subsection:orderbytask})
\item Does looping help parallelized layers? (\S \ref{subsection:loopedparallel})
\item Which variants harm performance the least? (\S \ref{subsection:leastharmful})
\end{enumerate} 
To answer these questions we perform a series of experiments on \emph{pretrained} LLMs.  These include experimenting with variations on the
standard transformer execution strategy, and measuring the impact of these variations on the models' performance across a variety of benchmarks
for both decoder-only (Llama) and encoder-only (BERT) models.  Note that our experiments never involve finetuning or otherwise adjusting the
models' parameters (with the caveat that the GLUE evaluation standard procedure includes a finetuning step for our BERT-Large
model)

\def\marcheight{0.15\textheight}
\def\marcwidth{0.15\textwidth}
\def\llamawidth{0.40\textwidth}

\begin{figure*}[t]
  \begin{subfigure}[t]{.1\textwidth}
    \includegraphics[height=\marcheight]{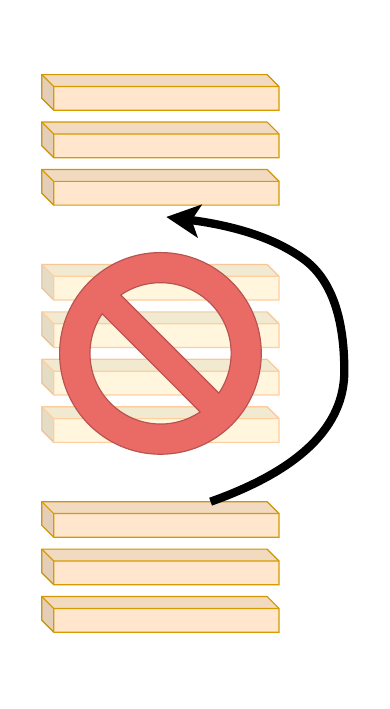}
    \caption{Skip}
    \label{subfigure:skip}
  \end{subfigure}
  \hspace{0pt}
  \begin{subfigure}[t]{.18\textwidth}
    \hspace{10pt}
    \includegraphics[height=\marcheight]{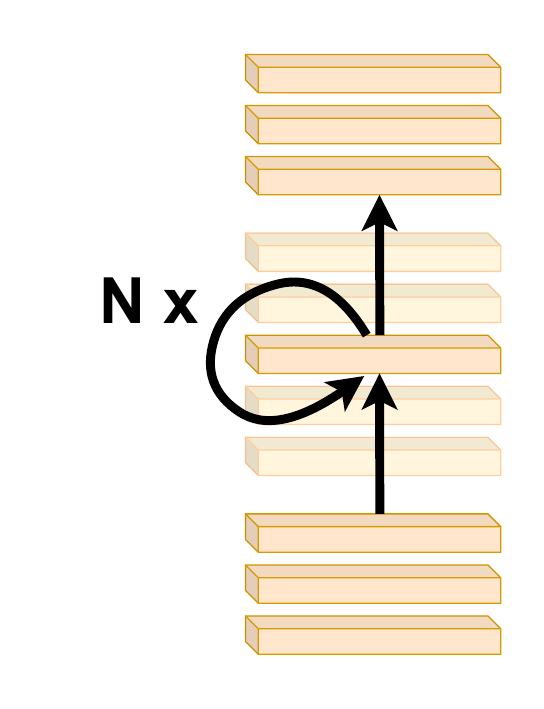}
    \caption{Middle Repeat}
    \label{subfigure:singlerepeat}
  \end{subfigure}
  \hspace{0pt}
  \begin{subfigure}[t]{0.12\textwidth}
    \includegraphics[height=\marcheight]{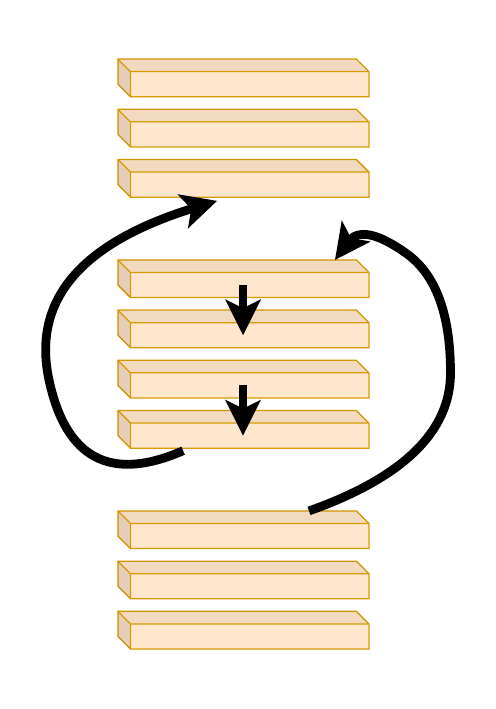}
    \caption{Reverse}
    \label{subfigure:reverse}
  \end{subfigure}
  \hspace{0pt}
  \begin{subfigure}[t]{0.24\textwidth}
    \includegraphics[height=\marcheight]{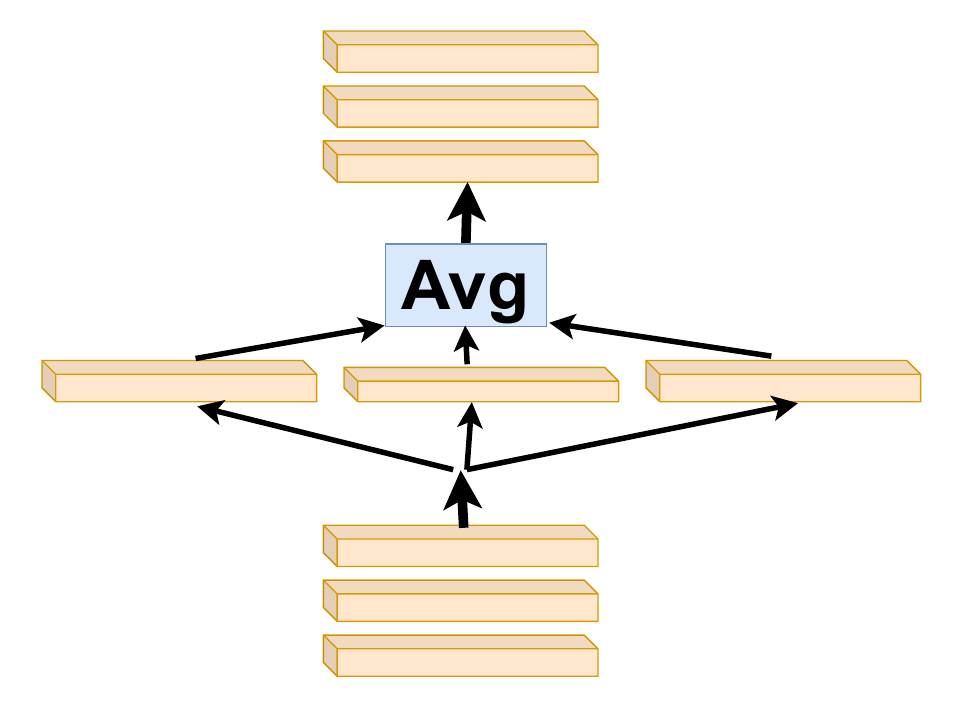}
    \caption{Parallel}
    \label{subfigure:parallel}
  \end{subfigure}
  \hspace{0pt}
  \begin{subfigure}[t]{0.2\textwidth}
    \includegraphics[height=\marcheight]{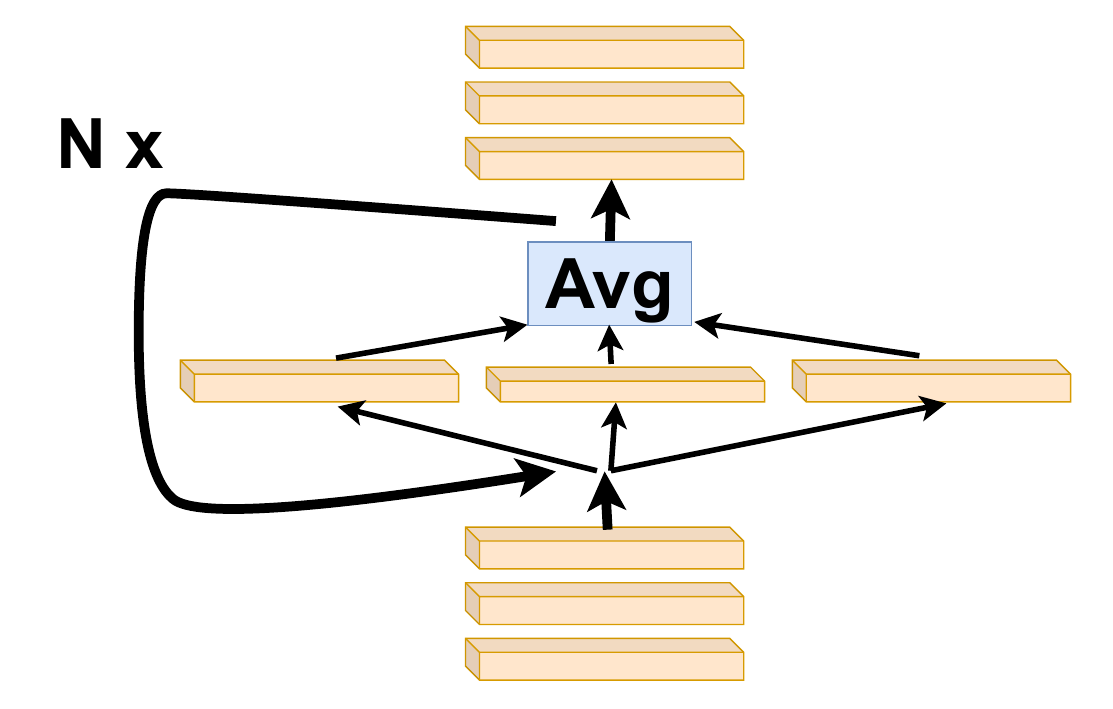}
    \caption{Looped Parallel}
    \label{subfigure:loopedparallel}
  \end{subfigure}
  \caption{Different execution strategies.}
  \vspace{-4mm}
\end{figure*}

\section{Models and Benchmarks}

Our experiments are primarily on two transformer models: Llama2 \cite{touvron2023llama}, and on BERT-Large \cite{devlin2019bert}.  (However, we
also include results for Mistral-7B \cite{jiang2023mistral7b} and Pythia-6.9B \cite{biderman2023pythiasuiteanalyzinglarge} in Appendix
\ref{subsec:othermodels} that support the generalization of our results.)  Llama2 is \emph{decoder-only}.  We focus on Llama2-7B, which has 7
billion parameters and 32 layers (each layer having 202 million parameters), but also include some scaling experiments with the 13B (40 layers)
and 70B (80 layers) models.  BERT is \emph{encoder-only} with 24 layers and 340 million parameters.  We used the standard pretrained checkpoints
for these models.  In all our experiments the models are frozen: we never modified the parameters of these models through fine-tuning or other
methods, with the exception of the BERT evaluation, which includes a standard fine-tuning step.

We used standard benchmarks for both \emph{decoder-only} LLMs (for Llama2) and for \emph{encoder-only} LLMs (for BERT).  For Llama2, we use {\bf
  ARC} (science exam questions) \cite{Clark2018ThinkYH}, {\bf HellaSwag} (commonsense) \cite{zellers2019hellaswag}, {\bf GSM8K} (Math Word
Problems) \cite{cobbe2021training}, {\bf WinoGrande} (Winograd Schema Challenge) \cite{sakaguchi2019winogrande}, and {\bf LAMBADA} (word
prediction) \cite{paperno2016lambada}.  This last, LAMBADA, measures perplexity and is closest to the raw token-prediction used during training.
For Llama2, we include the \emph{normalized median} of the benchmarks, where we scale each benchmark with 0 being the performance of random (or
max-class) guessing and 1 being the performance of the full Llama2 model.
For BERT, we used tasks from the GLUE benchmark \cite{wang-etal-2018-glue} and followed their evaluation protocol, including reporting the
\emph{unnormalized average} of the benchmarks.  Note that standard BERT evaluation includes a fine-tuning step \cite{devlin2019bert}, so our
BERT model has a chance to adapt to the new configuration.  Therefore, we also include results from an evaluation where an additional output
layer can adapt, but the model itself is frozen.  These results are in Appendix \ref{subsec:figures}, and more details of the GLUE benchmark are
given in Appendix \ref{subsec:benchmarks}.

\vspace{-2mm}
\section{Experiments}

The original motivation behind our experiments came from the question of whether multiple layers could be somehow be merged into a single
(possibly larger) layer.  (Such merging could potentially be automated \cite{akiba2024evolutionary}.)  We hypothesized, perhaps because of the
use of residual connections during training, that the middle layers of a neural network may use a common representation space.  (This is not the
case for standard multi-layer perceptrons, where there is nothing to encourage a common representation or permutational consistency across
layers.)  The possibility of layers sharing a common representation has downstream implications for conditional computation
(e.g. \cite{pagliardini2024denseformer}) or for dynamically inserting new knowledge into pretrained transformer models.

\subsection{\bf Do Layers “Speak the Same Language”?}
\label{subsection:samelang}

\begin{figure*}[ht]
  \centering
  \includegraphics[width=1.0\textwidth]{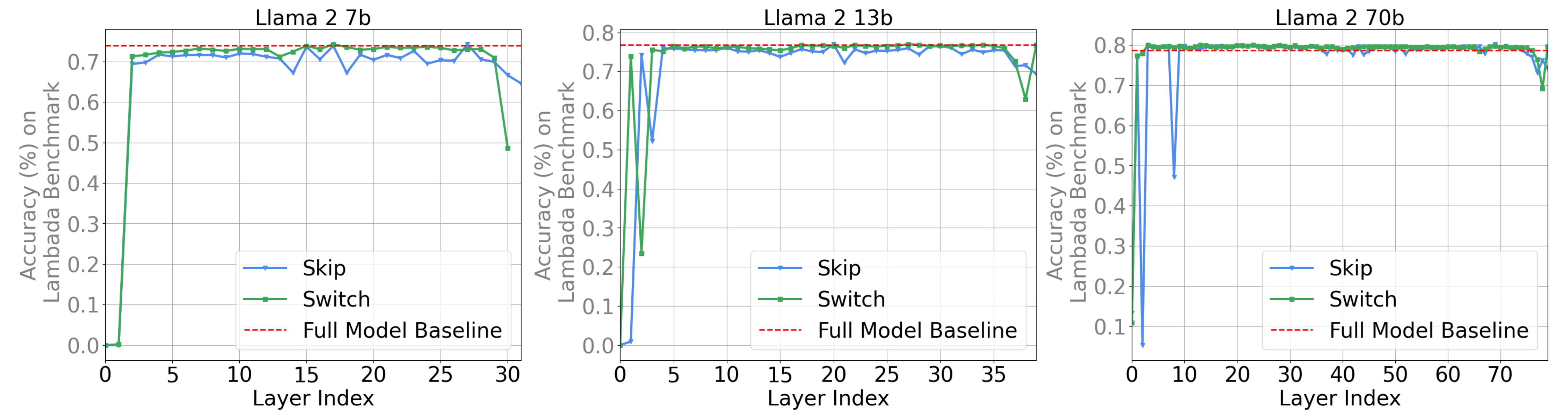}
  \label{figure:skipswitch}
  \vspace{-4mm}
  \caption{Results for Open-LAMBADA from \emph{skipping} layer $N$ (blue), and from \emph{switching} layer $N$ with $N+1$ (green) of Llama2-7B. Skipping early layers has a catastrophic effect, while the model is much more robust to skipping middle layers.}
\end{figure*}

To answer whether different layers have a shared representation space, we test whether transformers are robust to skipping specific layers or
switching the order of neighboring layers.  For example, in Llama2-7B, layer 6 normally expects the output from layer 5.  Would layer 6 behave
catastrophically if it were given layer 4's output instead?  In Figure \ref{figure:skipswitch}, we see that, with the important exception of the
first and last few layers, Llama2-7B's layers are fairly robust to skipping or even switching layers (e.g., feeding layer 4's output to layer 6,
then sending layer 6's output to layer 5, then to layer 7).


\begin{figure*}[ht]
  \centering
  \includegraphics[width=1\textwidth]{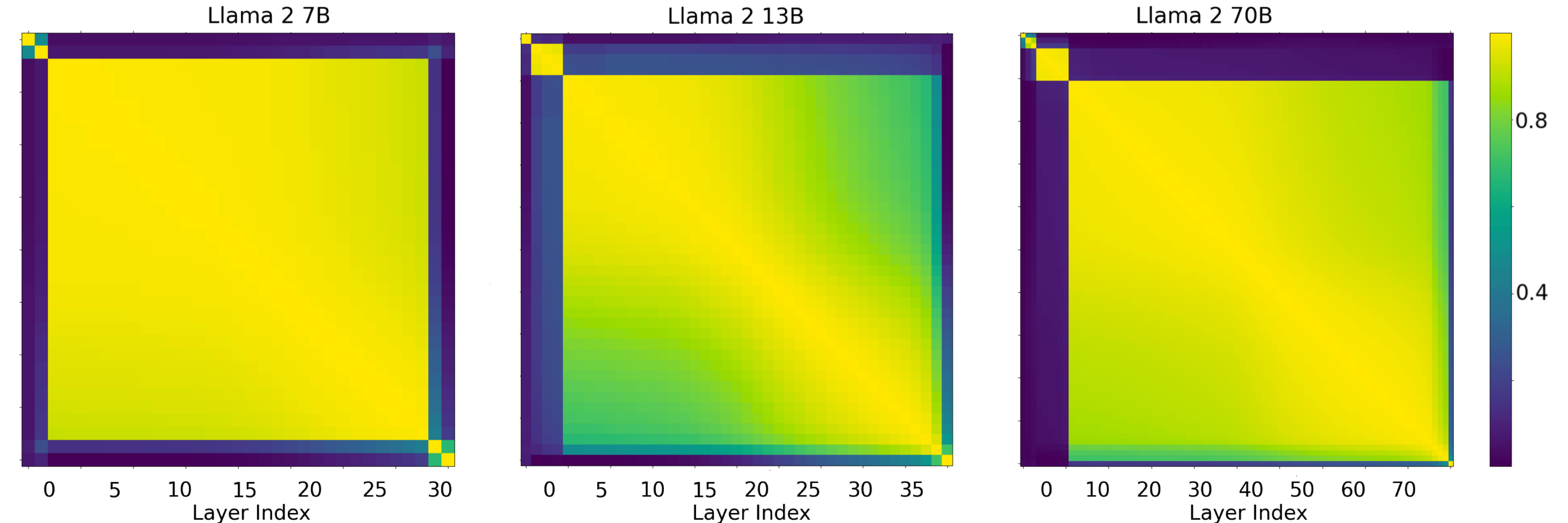}
  \caption{Avg.\ cosine similarity between the hidden states of all 32 layers of Llama2-7B (top) and all 40 layers of Llama2-13B.}
  \label{figure:llama2_suite_similarity}
  \label{figure:cossim}
  \vspace{-4mm}
\end{figure*}

This experiment would suggest that the middle layers 1. share a representation space and 2. have a separate representation space from the
``outer'' (first and last few) layers.  To further test this hypothesis, following previous work \cite{friedman2023comparing,
  kornblith2019similarity, simoulin-crabbe-2021-many, godey2024anisotropy, xue2023study}, we measured the average cosine similarity between the
activations of hidden states of different layers of our models (Llama2-7B, Llama2-13B, and BERT-Large) across our benchmarks.
In Figure \ref{figure:cossim}, we show that this consistency holds among all the middle layers.  For example, the activation in the fourth layer
from the bottom has a high similarity to the fourth layer from the top.  For the 40 layers of Llama2-13B, we see that the layers form four or
five distinct similarity groups: Layer 0, layers 1-3, the middle layers, then the final layer or two.

This suggests that the model may have three distinct representation spaces for the ``beginning'', ``middle'', and ``ending'' layers.  Note that
in the 13B model, the number of ``beginning layers'' is 3 while the 7b is 2, the ``ending layers'' is 1 or 2 and 7b is clearly 2.  So the number
of ``beginning layers'' seems to grow as the total number of layers increases.  (In Appendix \ref{subsec:scaling} we further show that these
three classes are consistent across different model scales, with the beginning and middle layers growing proportionally to the total number of
layers.)  Also note that a high cosine similarity \emph{may} suggest a shared representation space, but a low similarity is more indicative that
the spaces are \emph{not} shared.  However, the fact that the matrix for Llama2-7B in Figure \ref{figure:cossim} aligns neatly with the
performance shown in Figure \ref{figure:skipswitch} is stronger evidence that the \emph{semantics} of the representation space is actually
shared, at least for the middle layers.  Based on this, we answer this subsection's question with:

{\bf Yes, the middle layers seem to share a common representation space.}

\subsection{\bf Are All the Layers Necessary?}
\label{subsection:necessary}
\begin{figure}[ht]
  \centering
  \includegraphics[width=\llamawidth]{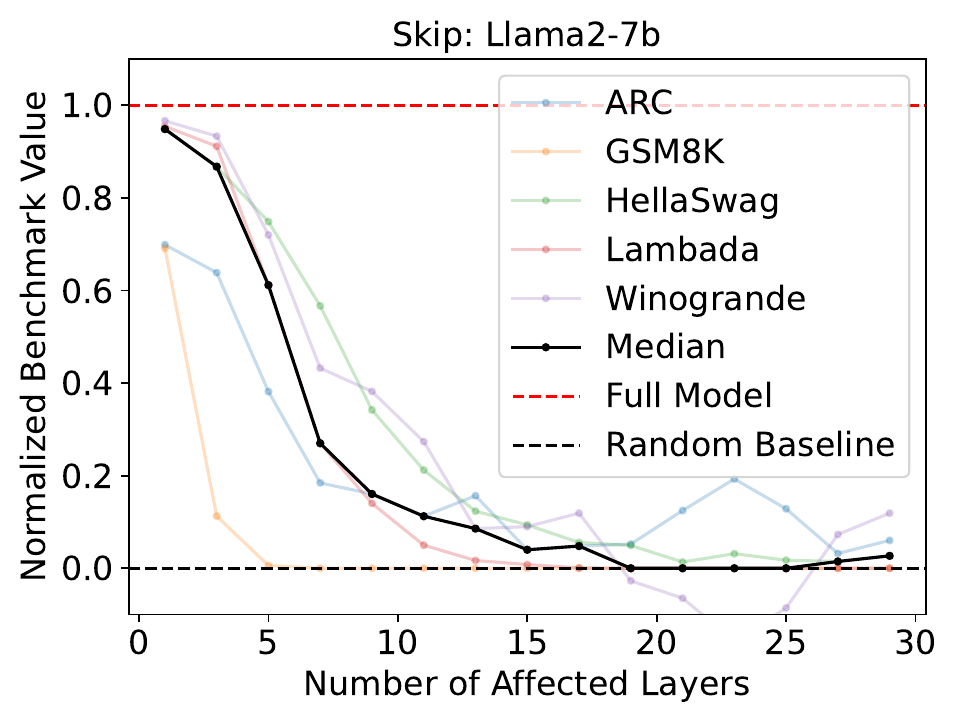}
  \includegraphics[width=\llamawidth]{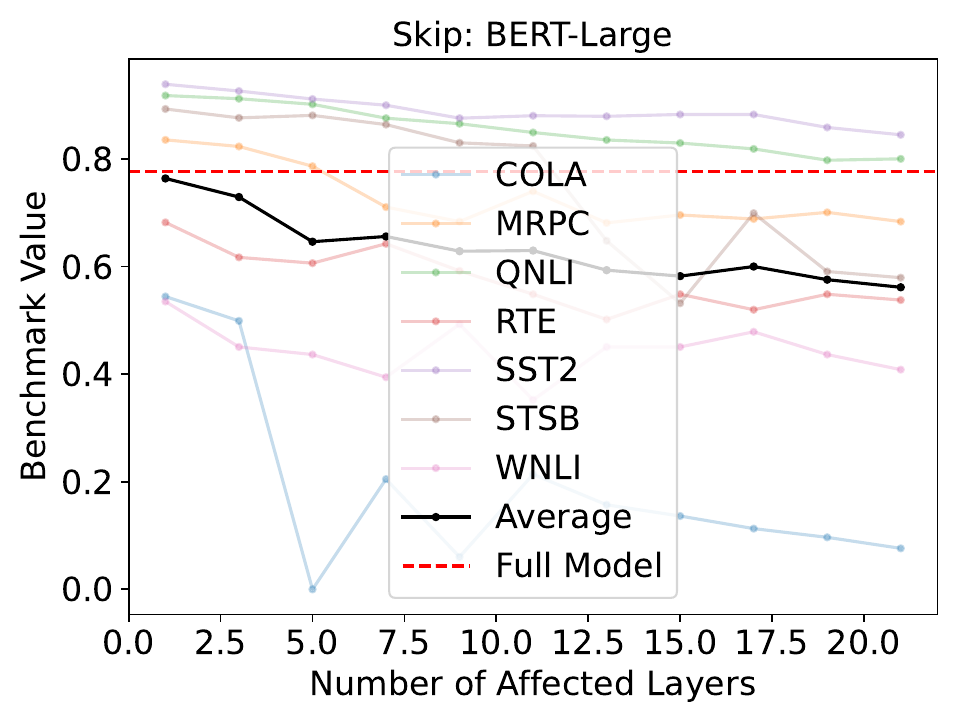}
  \caption{Top: Skipping layers N to 32-N for Llama2-7B, normalized per benchmark (median). Bottom: Skipping layers N to 24-N for BERT, with unnormalized average.}
  \label{figure:skip}
  \vspace{-4mm}
\end{figure}

To further test whether the reorientation space for middle layers is truly shared (in addition to having close cosine similarity), we experiment
with skipping layers.  That is, we send the output of the $N$th layer directly into the input of layer $N+M$ (where $M > 1$), thereby
``skipping'' $M-1$ layers, as illustrated in Figure \ref{subfigure:skip}.
Recall that we perform no fine-tuning during our experiments.  Our experiments are to see if layer $N+M$ can make sense of activations from
layer $N$, though it was trained only on inputs from layer $N+M-1$.  For this (and related) experiments, we execute the first and last $N-1$
layers as normal, skipping (or later modifying) layers $N+1$ through $T-N$, where $T$ is the total number of layers in the model.
Figure \ref{figure:skip} shows that performance for many of our benchmarks has graceful degradation for both Llama2-7B and BERT-Large.  (Note
that the number of layers skipped is inversely proportional to $N$, so the plot goes from few skipped layers to many skipped layers when read
from left to right.)  This result suggests that the answer to whether all the layers are necessary is:

{\bf No, at least a few middle layers can be dropped without catastrophic failure}.

\added{
In Appendix~\ref{sec:app:skip_at_scale}, we analyze the layer skipping behavior across model sizes, revealing a surprisingly uniform pattern in the importance of middle layer partitions. Furthermore, in Appendix~\ref{sec:app:skip_w_finetune}, we demonstrate that fine-tuning can enhance performance when skipping fewer layers but becomes harmful when skipping too many.
}

\subsection{\bf Are Middle Layers All Doing the Same Thing?}
\label{subsection:samething}

If the middle layers share a common representation space, does this mean that these layers are redundant?  To test this, we reran the ``Skip''
experiments from the previous subsection, but instead of skipping the middle layers, we replaced their weights with those of the center layer,
effectively looping on this layer for $T-2N+1$ times, where $T$ is the total number of layers (32 for Llama2-7B, 24 for BERT-Large).  (See
illustration in Figure \ref{subfigure:singlerepeat}.)
\begin{figure}[ht]
  \centering
  \includegraphics[width=\llamawidth]{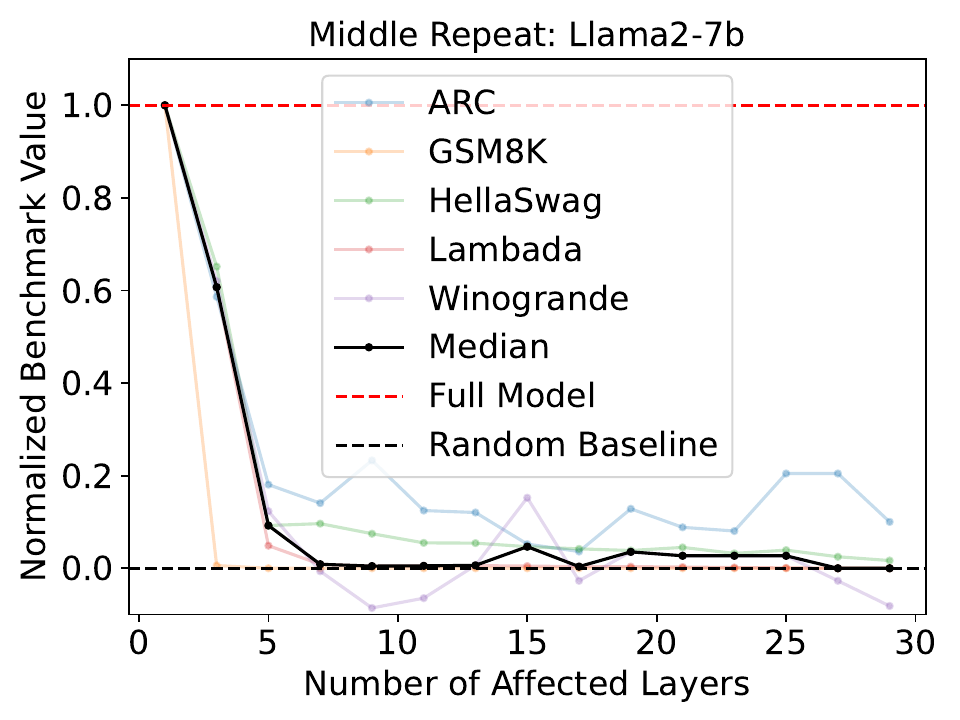}
  \includegraphics[width=\llamawidth]{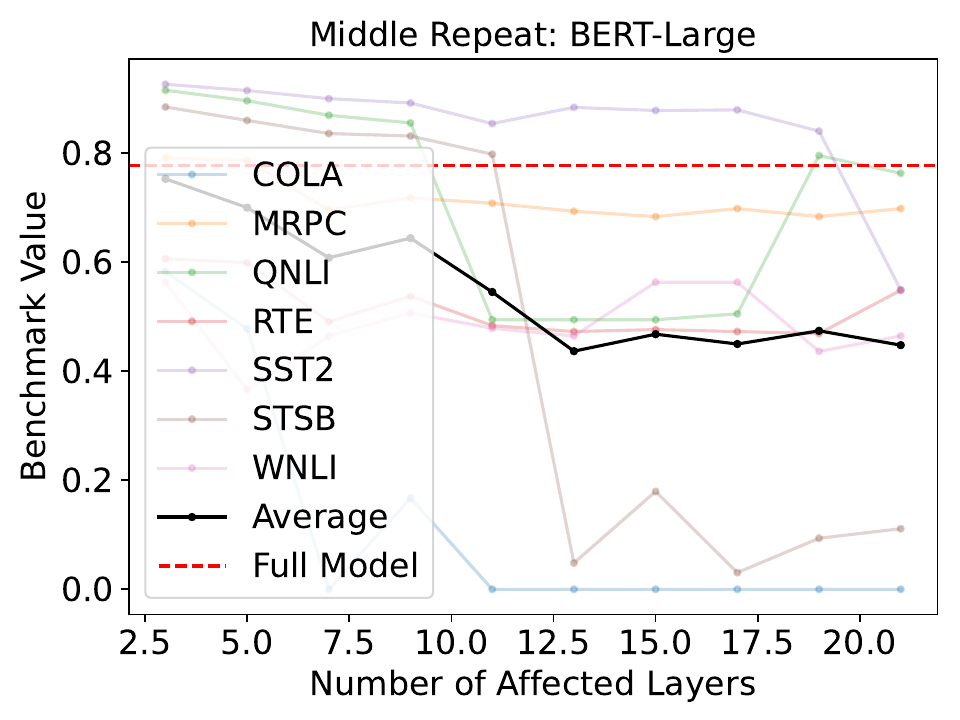}
  \caption{Replacing $M$ middle layers with the center layer (16 for Llama, 12 for BERT) for Llama2-7B (top, normalized benchmarks).  and BERT (unnormalized average).}
  \label{figure:repeated_middle}
  \vspace{-4mm}
\end{figure}

In Figure \ref{figure:repeated_middle}, we see that the benchmarks quickly decay as the number of replaced layers increases, and Figure
\ref{figure:allthethings} shows that this variation is the most catastrophic of all we tried, significantly worse than just skipping
layers\footnote{In Appendix \ref{subsec:whyrepeatworse}, we further explore why skipping is better than recycling the center-most layer.}.
Therefore, we answer our question with:

{\bf No, sharing weights among middle layers is catastrophic, indicating that the middle layers are performing different functions.}

\subsection{\bf Does the Layer Order Matter?}
\label{subsection:order}

The previous experiments suggest that middle layers share a representation space but perform different operations on this space.  Another
question is how much the order of these function matters.  We performed two sets of experiments to test this.  First, is running the middle
layers in reverse order from how they were trained\footnote{Again, we emphasize that there is no fine-tuning, so the layers can't merely adapt
to the new order.}.  Specifically, we take the output of layer $T-N$ and send it into the input of $T-N-1$, then the output of this layer into
$T-N-2$ and so on down to layer $N$, then send the output of this layer to the last $T-N$ layers.  (See Figure \ref{subfigure:reverse}.)  In the
second variation we ran the middle layers in a random order (and averaged the results over 10 seeds).

\begin{figure}[ht]
  \centering
  \includegraphics[width=\llamawidth]{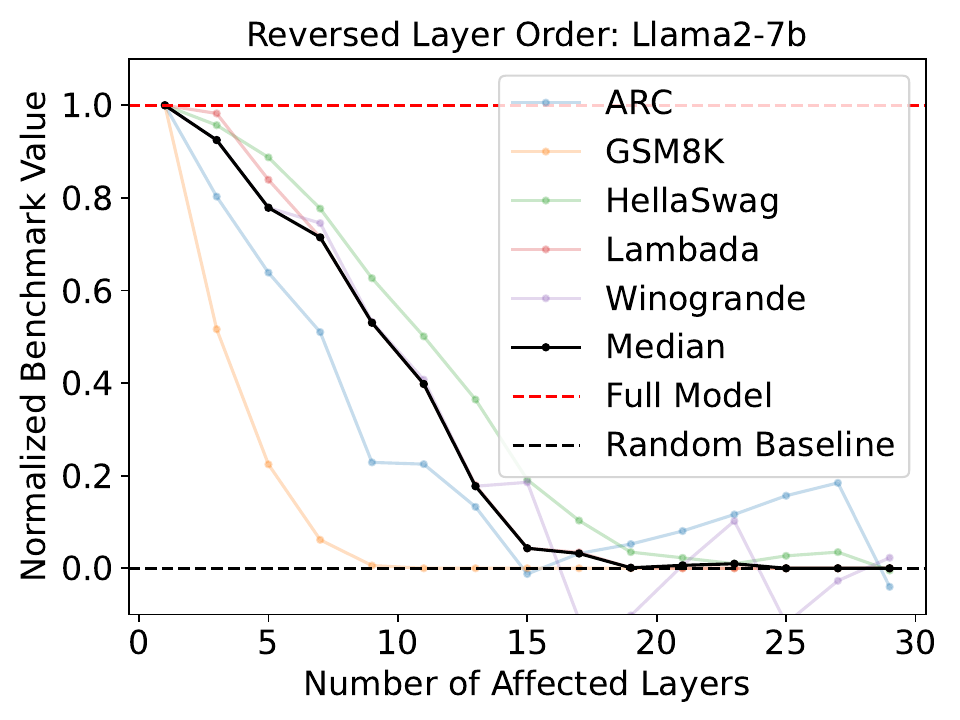}
  \includegraphics[width=\llamawidth]{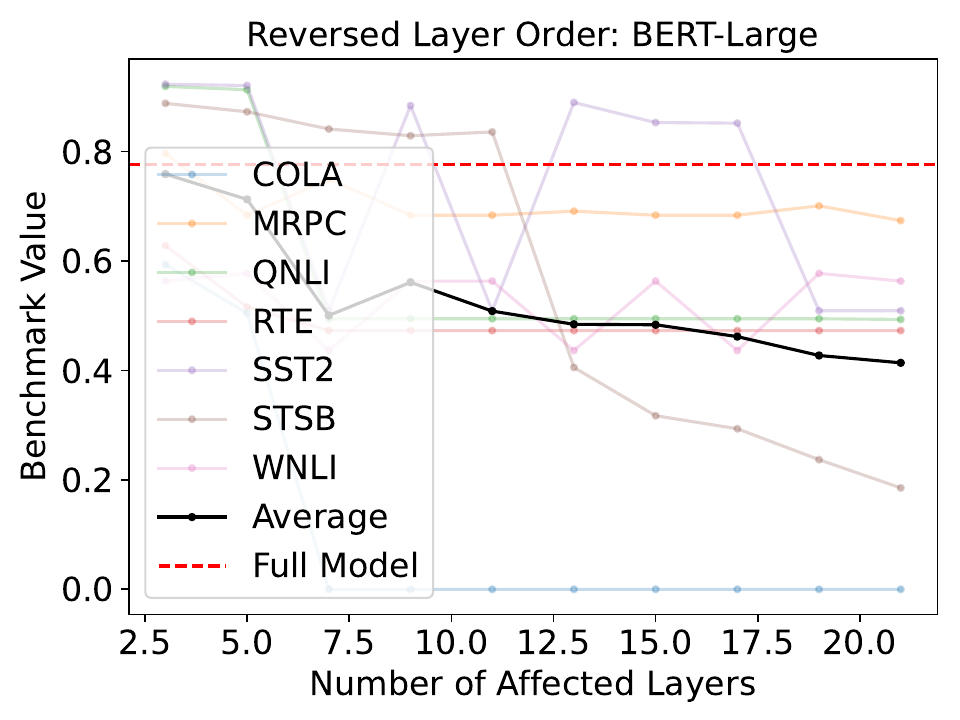}
  \caption{Top: Reversing $M$ middle layers for Llama2-7B, normalized across different Benchmarks. Bottom: Reversing layers for BERT-Large, unnormalized average.}
  \label{figure:reverse}
  \vspace{-4mm}
\end{figure}

\begin{figure}[ht]
  \centering
  \includegraphics[width=\llamawidth]{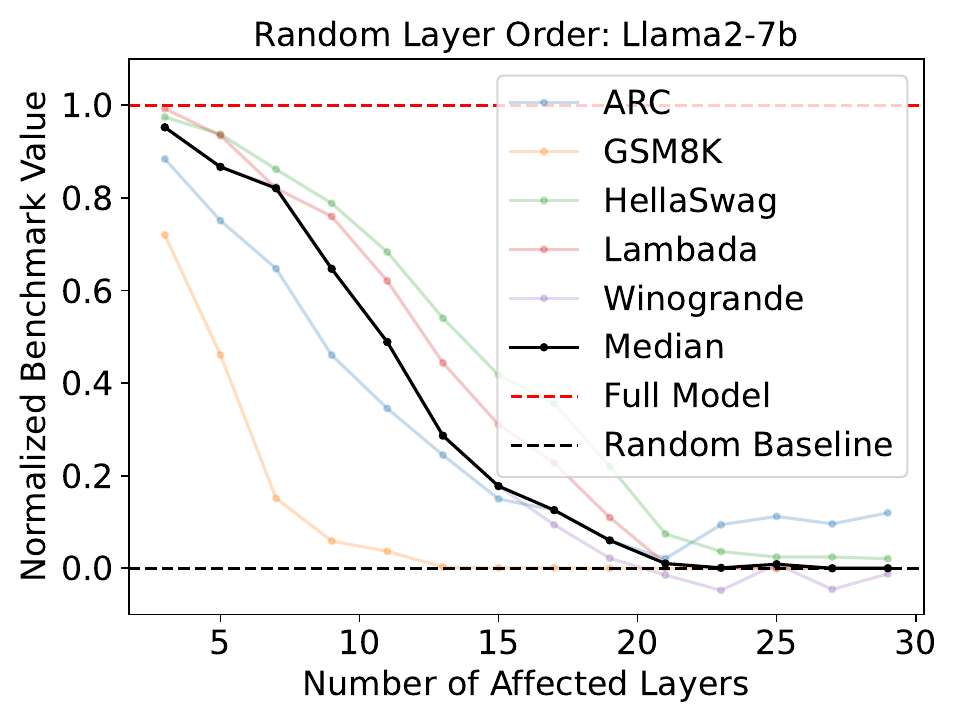}
  \includegraphics[width=\llamawidth]{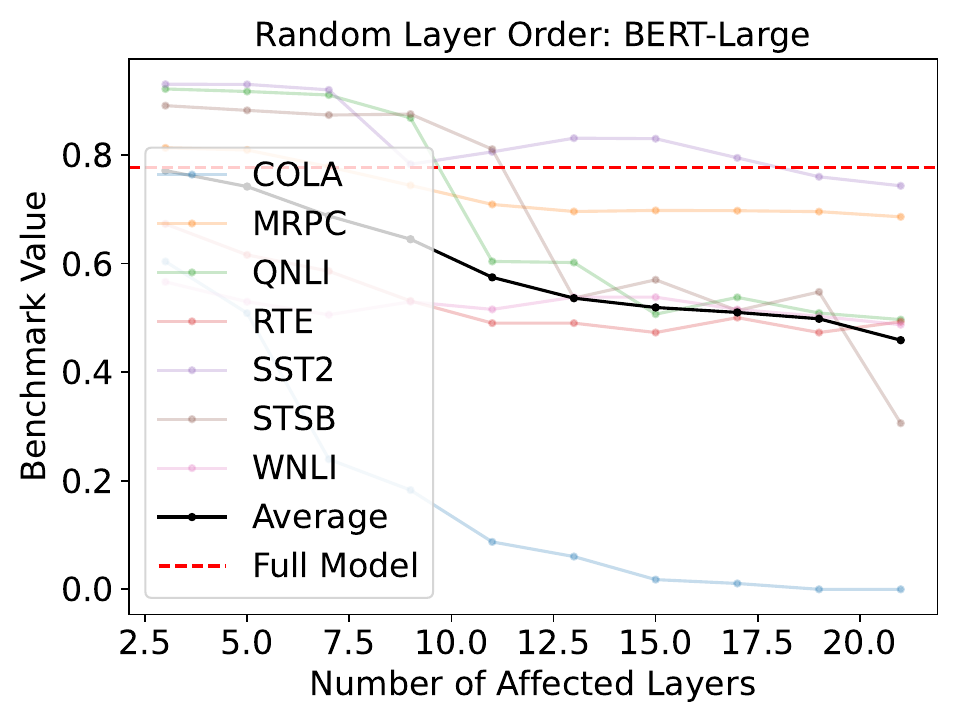}
  \caption{Randomizing layer order for $M$ middle layers for Llama2-7B (top) and BERT (bottom).  Each point is the average of 10 random seeds.}
  \label{figure:randorder}
  \vspace{-4mm}
\end{figure}

The results for Reversed and Random Order are shown in Figures \ref{figure:reverse} and \ref{figure:randorder}, respectively, each showing
graceful degradation.  Figure \ref{figure:allthethings} shows that both of these methods outperform Skipping the layers, suggesting that layers
are still able to contribute even when run on different input sources (i.e., different layers) from how they were trained.  Therefore, we answer
this subsection's question as:

{\bf Somewhat.  Both randomizing and reversing the middle layer order has graceful degradation.}

Interestingly, Random Order outperforms Reverse Order as can be seen more clearly in Figure \ref{figure:allthethings}.  One possible explanation
is that Reverse the exact opposite of the order in which the layers were trained.  So any random order will have at least as much consistency
(in that layer $i$ is after layer $j$, where $i > j$) as totally reversing the order.

\vspace{-2mm}
\subsection{\bf Can We Run the Layers in Parallel?}
\label{subsection:parallel}

If the presence of the layers (i.e., that they're not Skipped) is more important than the order in which they're executed, we may ask whether we
can run the layers \emph{independently} from an early input and merge their results, as illustrated in Figure \ref{subfigure:parallel}.  To
answer this, we ran an experiment where, instead of skipping layers $N$ through $T-N$, we ran these middle layers in parallel, then sent their
averaged result to the final $N$ layers.

\begin{figure}[ht]
  \centering
  \includegraphics[width=\llamawidth]{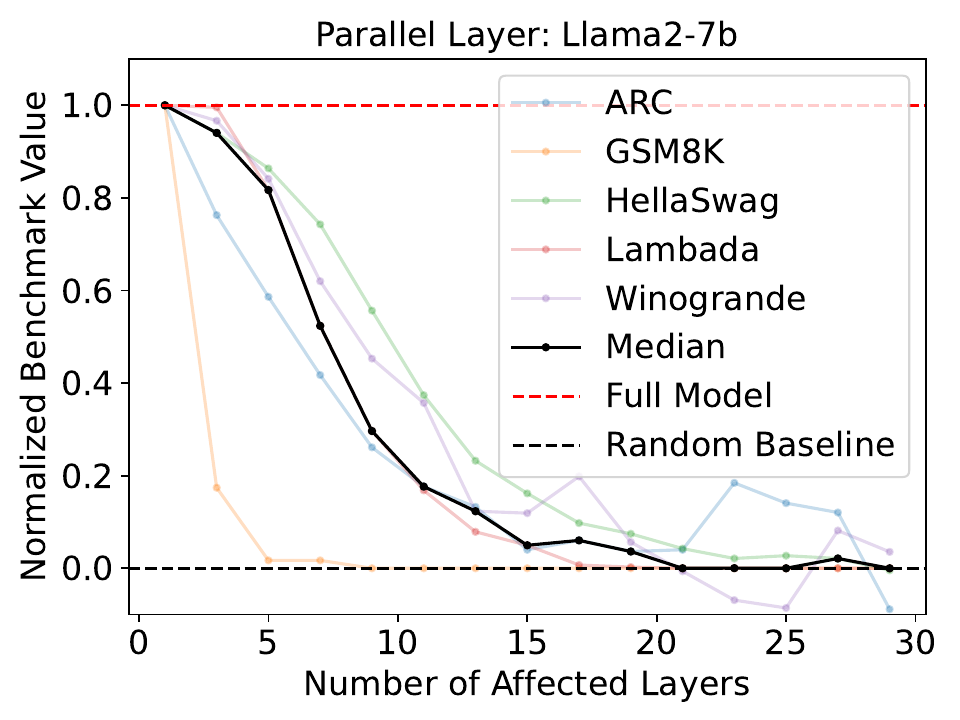}
  \includegraphics[width=\llamawidth]{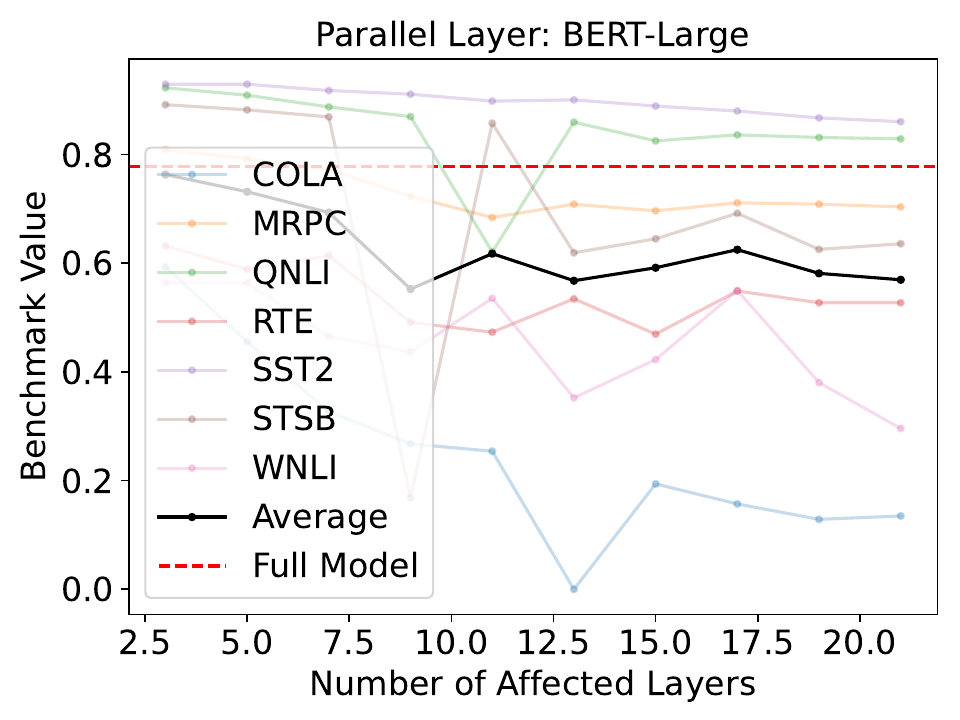}
  \caption{Running $M$ layers (Layers $(T-M)/2$ to $(T-M)/2$) in parallel for Llama2-7B (top) and BERT (bottom)}
  \label{figure:parallel}
  \vspace{-4mm}

\end{figure}

Figure \ref{figure:parallel} shows graceful degradation for all benchmarks except the GSM8K math word problems.  In Figure
\ref{figure:allthethings} this variation (``Parallel Layer'') outperforms skipping layers, but curiously does worse than running the layers in
reverse order.  In subsection \ref{subsection:orderbytask}, we further explore which benchmarks are most affected by our changes, so we answer
this subsection's questions with:

{\bf Yes, except for our math-heavy benchmarks.}

\subsection{\bf Does the Order Matter for Some Tasks More Than Others?}
\label{subsection:orderbytask}
Note that abstract (ARC) or mathematical (GSM8K) reasoning benchmarks have the steepest decline for most variants, including \emph{Reversed},
\emph{Skip}, and \emph{Parallel}.  One interpretation is that step-by-step reasoning tasks are more sensitive to layer order than ``semantic''
tasks like Winogrande or HellaSwag (Commonsense).  This is because reasoning involves both structure and semantics to perform well compared with
tasks like HellaSwag where semantics are enough to complete the task.  This would be consistent with the hypothesis that some degree of
order-dependent reasoning is happening within a single pass of the model.  In our Painter analogy, a semantic task would be analogous to
painting a collage, where ordering is less dependent, where a reasoning task might be more like painting a precise architectural scene.
Regardless of whether the analogy holds, we empirically conclude that:

{\bf Yes!  Mathematical and reasoning tasks are more order dependent than ``semantic'' tasks.}

In Appendix \ref{appendix:gsm8k} we show a specific example that indicates that errors for GSM8K may come from \emph{arithmetic} errors.

\subsection{\bf Does Looping Help Parallelized Layers?}
\label{subsection:loopedparallel}
Following the Painter analogy, it's conceivable that some layers only ``add'' to the painting when given the appropriate input.  For example,
the ``wheel'' painter will be more likely to draw some wheels if she sees the body of a car first.  In transformer terms, layers might only
contribute to a forward pass --as opposed to ``passing'' the input forward via the residual connection-- when given the appropriate input.  If
this is the case, then iterating the parallelized layer from the previous experiment should improve performance compared to a single execution
of the parallelized layer.  We test this by feeding the mean output of the parallelized layer back into the same layer for a fixed number of
iterations, as shown in Figure \ref{subfigure:loopedparallel}.

\begin{figure}[ht]
  \centering
  \includegraphics[width=\llamawidth]{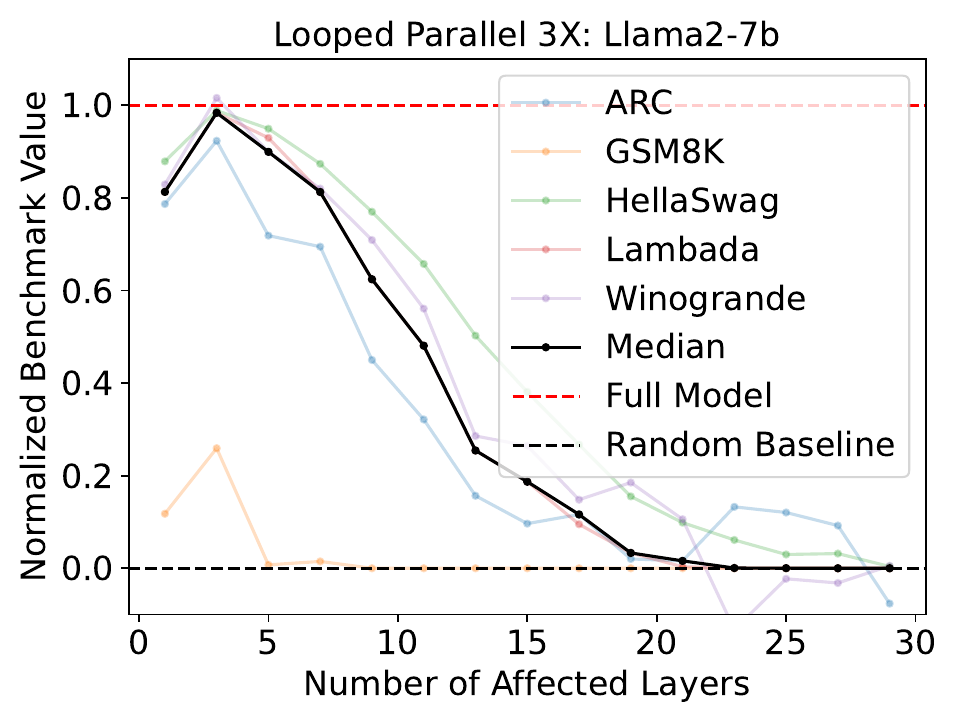}
  \includegraphics[width=\llamawidth]{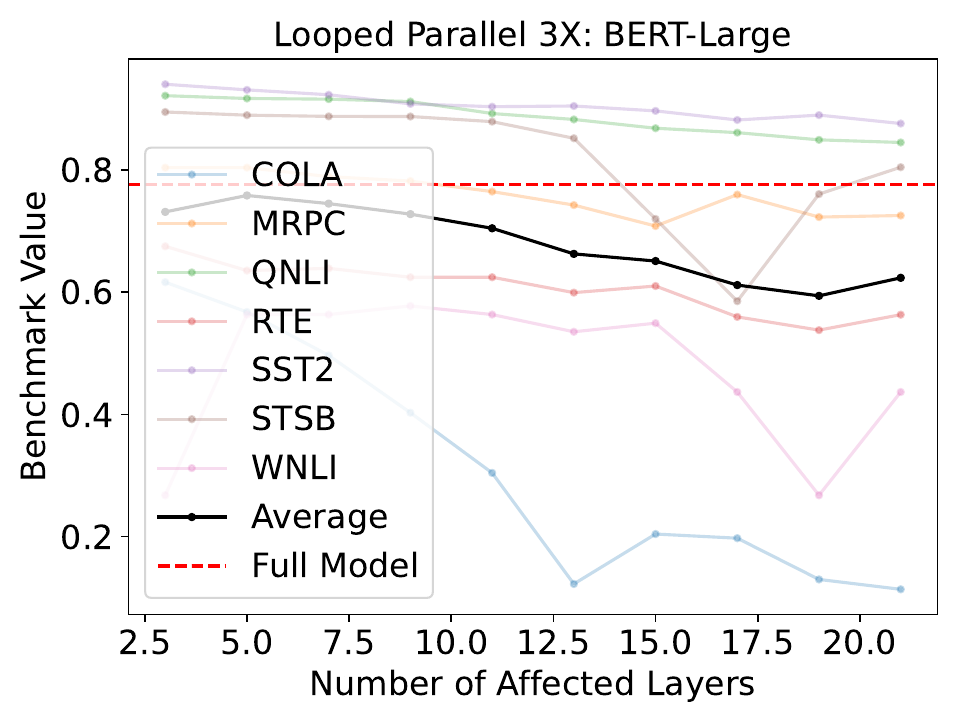}
  \caption{Running $M$ layers in parallel, looping 3 times for Llama2 (top) and BERT (bottom).}
  \label{figure:tripleloop}
  \vspace{-4mm}
\end{figure}

In Figure \ref{figure:tripleloop}, we show the results for looping the parallelized layer 3 times.  As can be seen in Figure
\ref{figure:allthethings}, this method (\emph{Looped Parallel 3X}) significantly improves on a single iteration (\emph{Parallel Layer}).  The
one exception is when the starting layer $N$ is 15 for Llama2-7B or 11 for BERT (the left-most cases for each, where only a single layer is
affected).  In this case, the \emph{ Looped Parallel 3X} model is equivalent to repeating only the middle layer 3 times, while the
\emph{Parallel Layer} for this point is equivalent to the full model.

\begin{figure}
  \centering
  \includegraphics[width=0.48\textwidth]{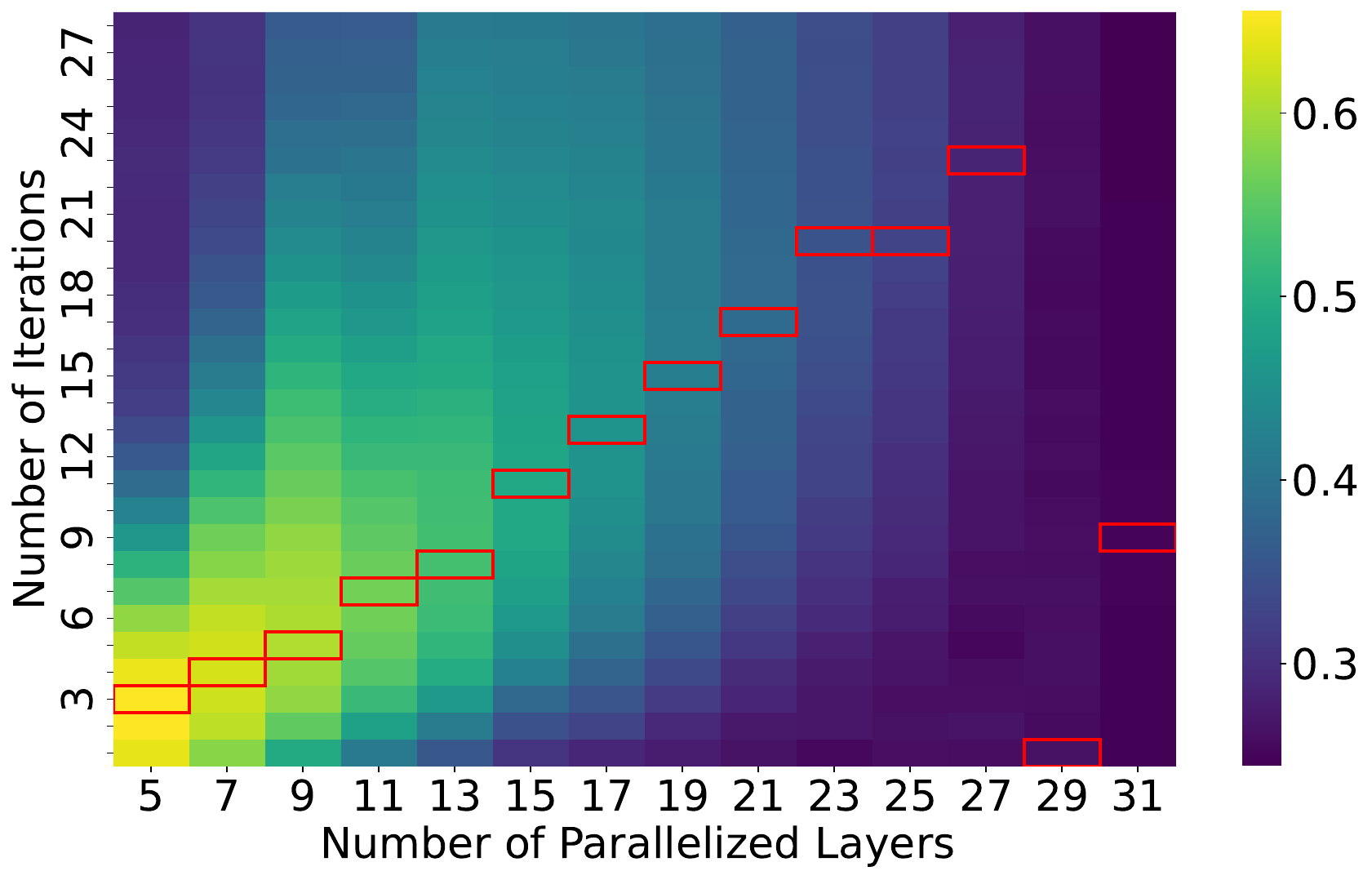}
  \label{figure:optimalloops}
  \caption{Looping parallelized layers of Llama2-7B, iterating from 1 to 28 times.  For each number of parallelized layers, the best iteration number is marked by a red box.}
  \vspace{-6mm}
\end{figure}

We also repeated the same experiment for different numbers of iterations.  In Figure \ref{figure:optimalloops}, we show performance for
Llama2-7B as a function of the number of parallelized layers $M$ and the number of iterations.  The highest performing loop iterations for each
$M$ is shown by a red box.  With the exception of $M=29$ and $M=31$ (parallelizing nearly all the layers), the optimal number of iterations is
roughly linearly proportional to the number of parallelized layers.  Therefore, we answer that:

{\bf Yes, with the optimal number of iterations proportional to the number of parallelized layers.}

\subsection{\bf Which Variants Are Least Harmful?}
\label{subsection:leastharmful}

\begin{figure}[ht]
  \centering
  \includegraphics[width=\llamawidth]{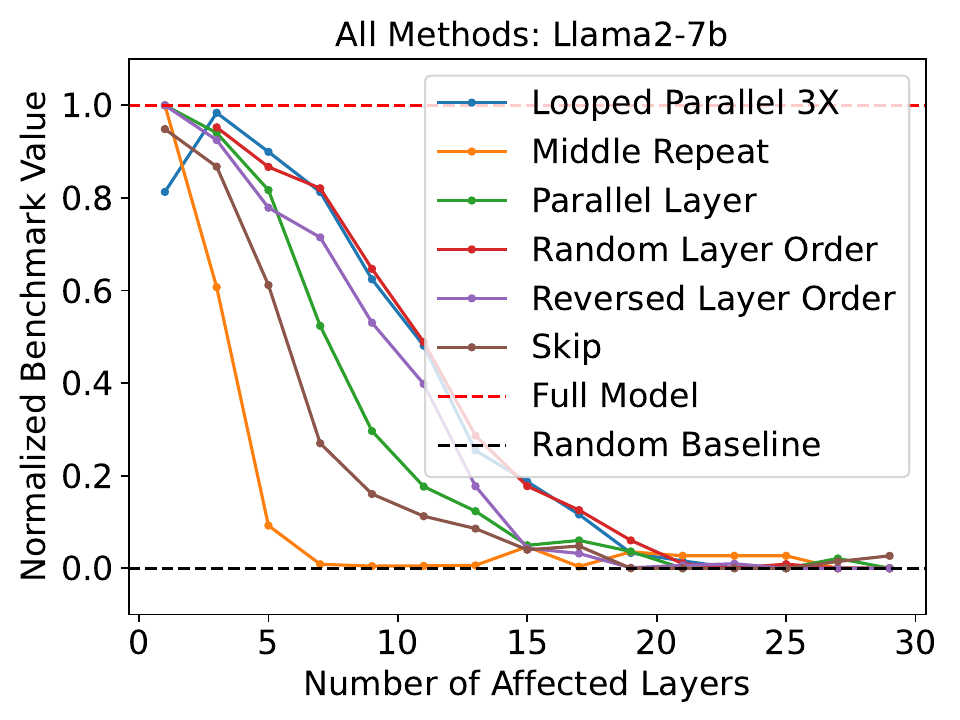}
  \includegraphics[width=\llamawidth]{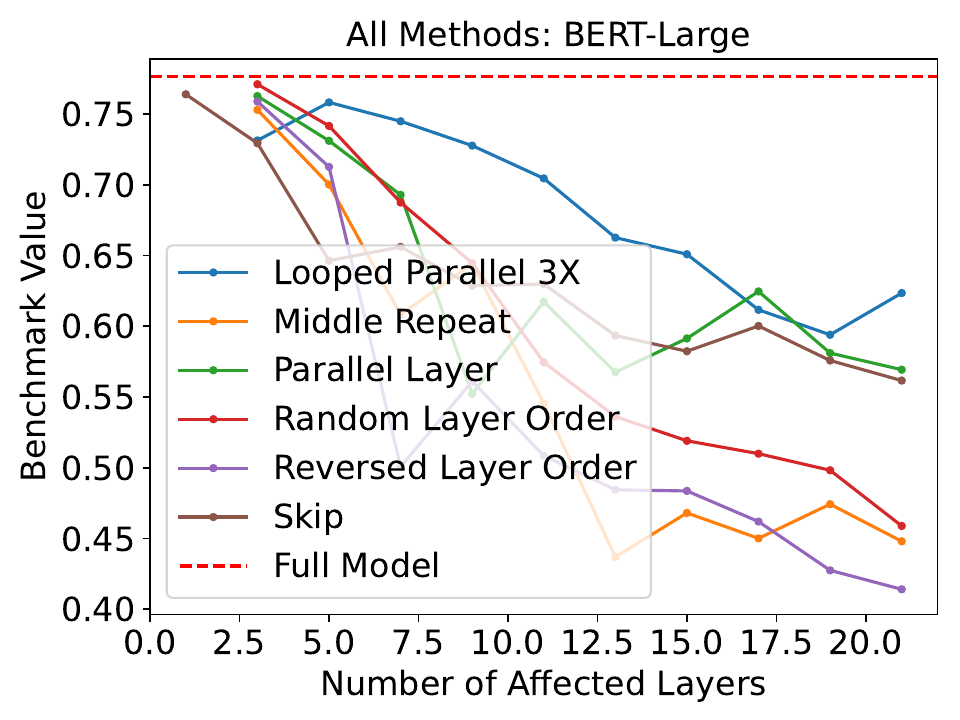}
  \caption{Average benchmark scores for different variations for Llama2-7B (top) and BERT-large (bottom).}
  \label{figure:allthethings}
  \vspace{-6mm}
\end{figure}

Finally, in Figure \ref{figure:allthethings} we compare all the different variants in our experiments on a single plot, showing the median (for
Llama2) or average (for BERT) performance over all the benchmarks.  Middle Repeat --replacing a period of middle layers with exactly the same
number of copies of the middlemost layer-- does worst by far, quickly degrading to random baseline performance.  On the other hand,
looped-parallel and random layer order have the shallowest degradation, with the former the best variant for both BERT and Llama2-7B.  So we
answer:

{\bf Repeating a single layer is worst.  Randomizing the layer order and looped-parallel do the least damage.}

These experiments generally show graceful degradation, but we still have the question of why the layers are somewhat robust to most of our
perturbations.  We offer a few suggestions in the Discussion section, but leave a full explanation for future work.

\section{Related Work}
A transformer layer contains a pair of multi-head attention (MHA) and feed-forward network (FFN), and almost all of the prior works focused on
finding a combination of them that works best, or reducing the parameter count in one way or another.  Our work offers an additional
perspective, in that we also investigate parallelizing and reusing layers.

\cite{kim2024shortened} showcased that pruning entire transformers layers can reduce latency without a considerable drop in performance. This is
in line with the findings in \cite{bhojanapalli2021understanding}. Also, both the works noted that the performance drop is substantial if we
drop the first few entire transformer layers. Hence there is an agreement that the first few transformers layers are crucial for performance.
One implication of this observation is that many of these layers would be carrying redundant information, and this was shown by
\cite{kim2024shortened} who removed these layers, and noticed the change in the PPL score. The authors then removed these layers in one-shot,
and retrained the model with LoRA to make up for the lost performance,

One aspect where \cite{bhojanapalli2021understanding} and \cite{kim2024shortened} observations differ though is the fine-grained
units. \cite{bhojanapalli2021understanding} observed that removing MLP layers have lesser impact on performance compared to removing an entire
transformer layer, whereas \cite{kim2024shortened} observed that this behavior is very much dependent on the size of the models.  They noted
that removing individual MHA and FFN modules results in better downstream task accuracy but worse PPL compared to removing entire transformer
layers when the model has more than 5B parameters.  For smaller models than 5B, layer-level pruning achieves superior results.
While \cite{kim2024shortened} did a successful job on pruning the models, the authors observed an (un)interesting side effect of the same.  The
pruned models perform worse when responding to factual questions or generating long responses.  The authors couldn't make up for the lost
performance on these tasks even after retraining the models, suggesting that while much of the information stored in these layers was
redundant, some parts of it were required for critical tasks e.g. factual Q\&A.

The experiments of ShortGPT \cite{men2024shortgpt} corroborate the findings of ShortenedLlama, exploiting the redundancy in LLMs to derive a
pruning technique.  Denseformer \cite{pagliardini2024denseformer} had similar findings where they found that modules even after applying DWA had
cosine similarity with original transformer modules, suggesting both that there is some redundant information flow, and that this can be
leveraged for sparsity.

More recently, \cite{freiberger2024layershuffleenhancingrobustnessvision} explores layer shuffling during training to enhance robustness of the
models, while \cite{dutta2024vtransacceleratingtransformercompression} proposes an algorithm that can be used for efficient pruning of
transformers.  \cite{lad2024remarkablerobustnessllmsstages} explores the robustness of transformer-based LLMs by deleting or swapping layers.
\cite{zou2024cqilinferencelatencyoptimization} focuses on efficient inference by splitting layers in groups, running them in parallel or
bypassing them.  On a similar note, \cite{flynn2024statshrinkingtransformerstraining} focuses on pruning transformers in different ways (entire
attention blocks, ffn, etc.).  Our work is more closely related to \cite{lad2024remarkablerobustnessllmsstages} and
\cite{flynn2024statshrinkingtransformerstraining} where the ablations involve frozen models.  We present a super set of such ablations for the
frozen transformer models.

\section{Discussion}

In this paper, we examined several questions raised by the Layers as Painters analogy.  Among our more interesting findings are: 1. There are
three distinct classes of layers (with Middle being the largest).  2. The middle layers have some degree of uniformity (but not redundancy).
And 3. Execution order matters more for math and reasoning tasks than semantic tasks. \added{We welcome future theoretical analysis of layer behaviors in transformer architectures based on our empirical findings.}

We leave a full explanation for why transformers are robust to our variations for future work.  One possible hypothesis is that the residual
connections during training are necessary for the layers to share a common representation.  It's already known that residual connections are
useful to help address the vanishing gradient problem \cite{he2015deep}, and that transformers trained without these connections perform worse
than without.  However, it would be interesting to rerun our variations on models without residuals, and see if our variations destroyed
whatever meager gains full non-residual models achieved.

We also plan to ``thaw'' our models and investigate if transformers take to adjust to the variations in the paper via fine-tuning.  If these models were fine-tuned with new architectures, the performance would probably be even better. 
It is worth noting that Parallel and Skip both have potentially lower latencies than the full model (assuming enough memory to execute the layers simultaneously).
For example, the latency for the Parallel Layer for Llama2-7B for N=8 should be about half that of normal Llama2-7B.
Though the aim of this paper is to better understand layers in transformer-based LLMs as opposed to introducing new models, our results suggest simple methods to easily trade accuracy for latency gains.
Our results also suggest that a routing mechanism for executing frozen layers may be used here, analogous to Switch Transformers \cite{fedus2022switch}.

\clearpage
\section*{Acknowledgements}
We would like to thank Owen He, who came up with the painter analogy after seeing some of our early results.  We would also like to thank Yujin Tang for providing valuable suggestions during the rebuttal process.

\bibliography{references}

\clearpage
\clearpage
\appendix
\section{Appendix}
\vspace{-4mm}

\label{sec:appendix}

\added{
\subsection{\bf Does skipping layers affect on the models changes across scales?}
\label{sec:app:skip_at_scale}
Emerging behavior of foundation models indicates different behaviors of models across sizes. An interesting question to ask is how does scaling affect the skipping intervention? We further launched the skip experiment for Llama2-13B and Llama2-70B to answer this question.
}

\begin{figure}[ht]
    \centering
    \includegraphics[width=\llamawidth]{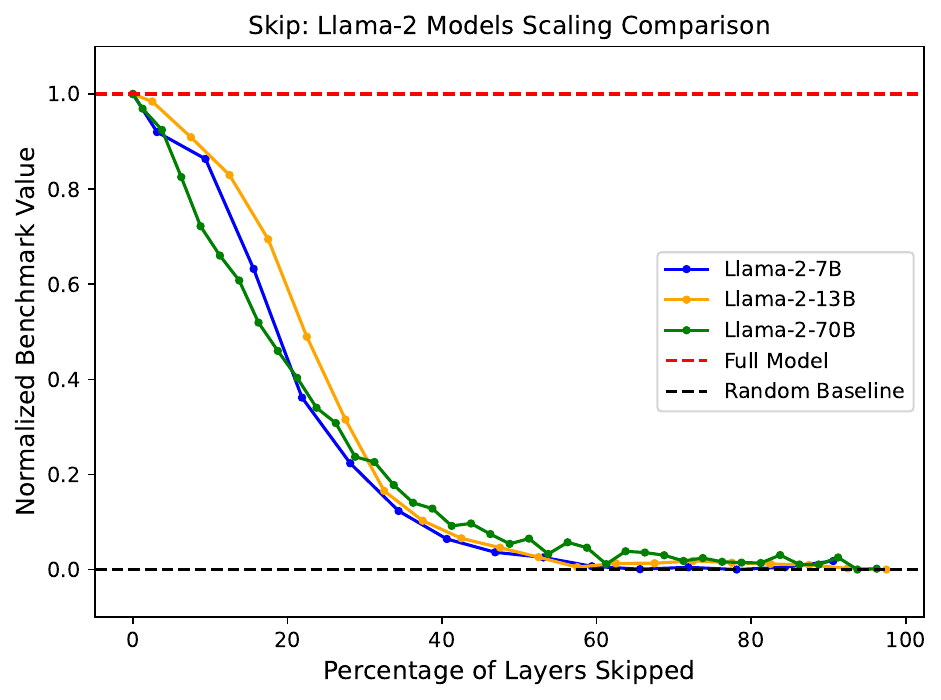}
    \caption{Skipping layers N to 32-N for Llama2-7B, 13B, and 70B.}
    \label{figure:skip_at_scale}
\end{figure}

\added{
We unified the number of skipping layers across different model sizes into the percentage of skipped layers to get Figure~\ref{figure:skip_at_scale}. Surprisingly, all models showed similar trends in their retained performance, So we answer:
}  

\added{
\bf{No, The affect remains consistent across scales, as middle layers partitions shows uniform importance regardless of model scale.}
}

\added{
\subsection{\bf How does fine-tuning affect the intervened model?}
\label{sec:app:skip_w_finetune}
Fine-tuning is a powerful technique for improving model performance on specific tasks. And How does it affect the model with layer skipping? To understand this, we conduct experiments using Llama2-7B on the ARC challenge. We fully fine-tuned models with different numbers of skipped layers for $1$ epochs using a learning rate of $5e-5$ and batch size of $64$ on the ARC training set.
}

\begin{figure}[ht]
    \centering
    \includegraphics[width=\llamawidth]{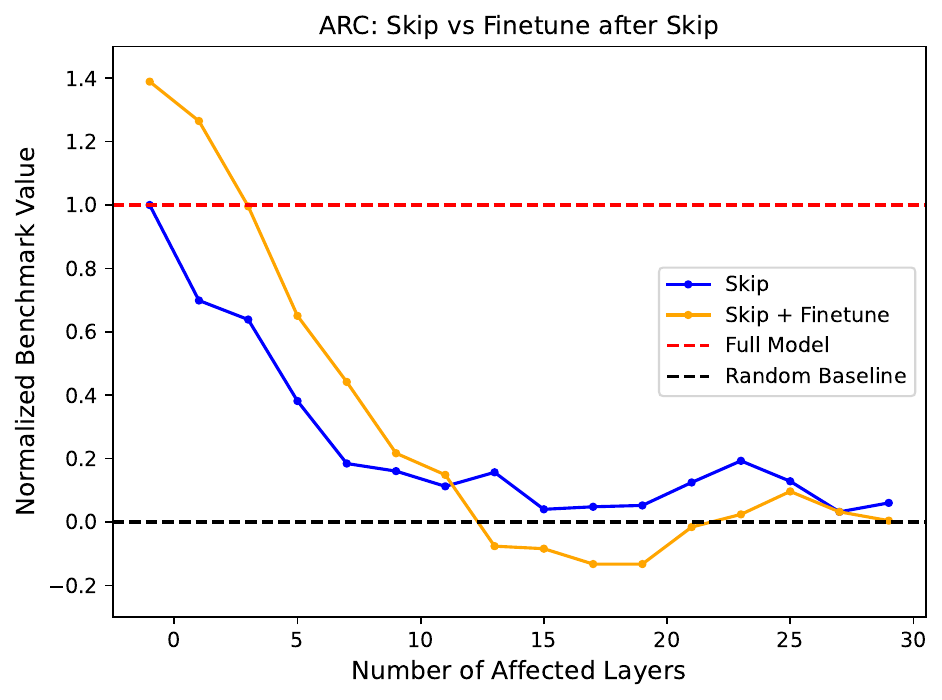}
    \caption{Comparison between directly skipping layers and fine-tuning the model after skipping layers on ARC challenge.}
    \label{figure:finetune}
\end{figure}

\added{
Figure~\ref{figure:finetune} reveals that fine-tuning can indeed improve model robustness when fewer than $30\%$ of layers are skipped, with these models showing slower performance degradation compared to their frozen counterparts. We also noted that fine-tuning hurts performance when more than $30\%$ of layers are skipped. Therefore:
}  

\added{
\bf{Fine-tuning benefits benign intervention but proves harmful for catastrophic intervention.}
}

\subsection{\bf How does scaling affect hidden state similarities?}
\label{subsec:scaling}

\begin{figure*}[ht]
  \centering
  \includegraphics[width=0.9\textwidth]{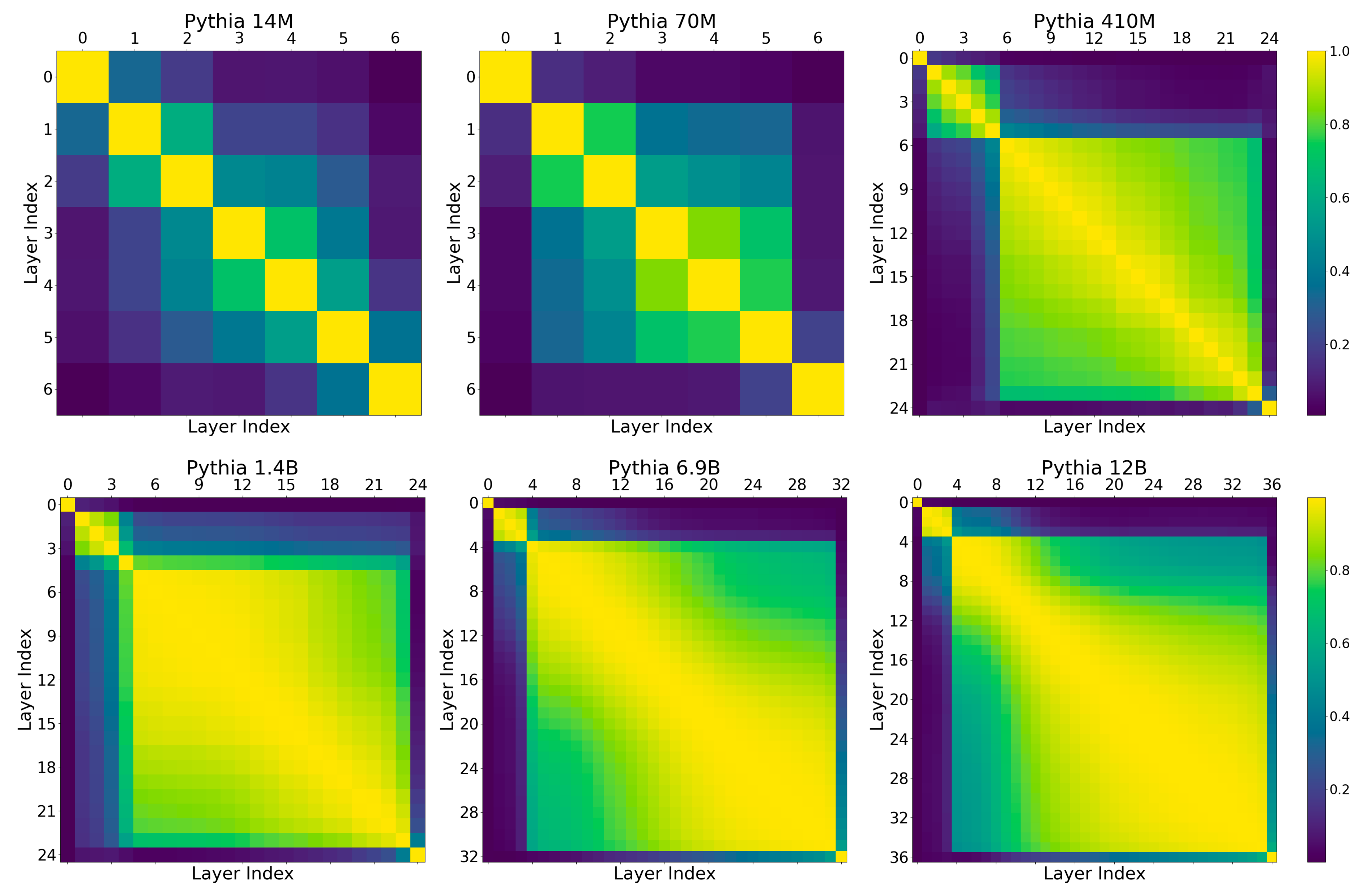}
  \caption{Average cosine similarity between the hidden states of all layers of pythia model in different size.}
  \label{figure:pythia_suite_similarity}
  \label{figure:scaling_cossim}
\end{figure*}

To reveal how the property of ``sharing similar embeddings'' evolves while increasing model size, we computed the average similarity matrices
for Pythia models (\cite{biderman2023pythia}) ranging in size from 14M to 12B parameters, utilizing 100 data samples from the LAMBADA token
prediction benchmark \cite{paperno2016lambada}.  Figure \ref{figure:pythia_suite_similarity} shows even the tiny 70M model has a clear boundary
for its beginning, middle, ending layers.  Interestingly, the \emph{proportion} of beginning layers decreases with an increase in model size but
the number of ending layer remains consistently at a single layer across all Pythia models, regardless of size.  We therefore answer our
question with:

{\bf The number of beginning and middle layers grows in proportion to the total number of layers, but the ``ending'' layers remains fixed at a
  single layer.}

\subsection{\bf Why is repeating a layer is so much worse than skipping it?}
\label{subsec:whyrepeatworse}

\begin{figure*}[ht]
  \centering
  \includegraphics[width=0.85
\textwidth]{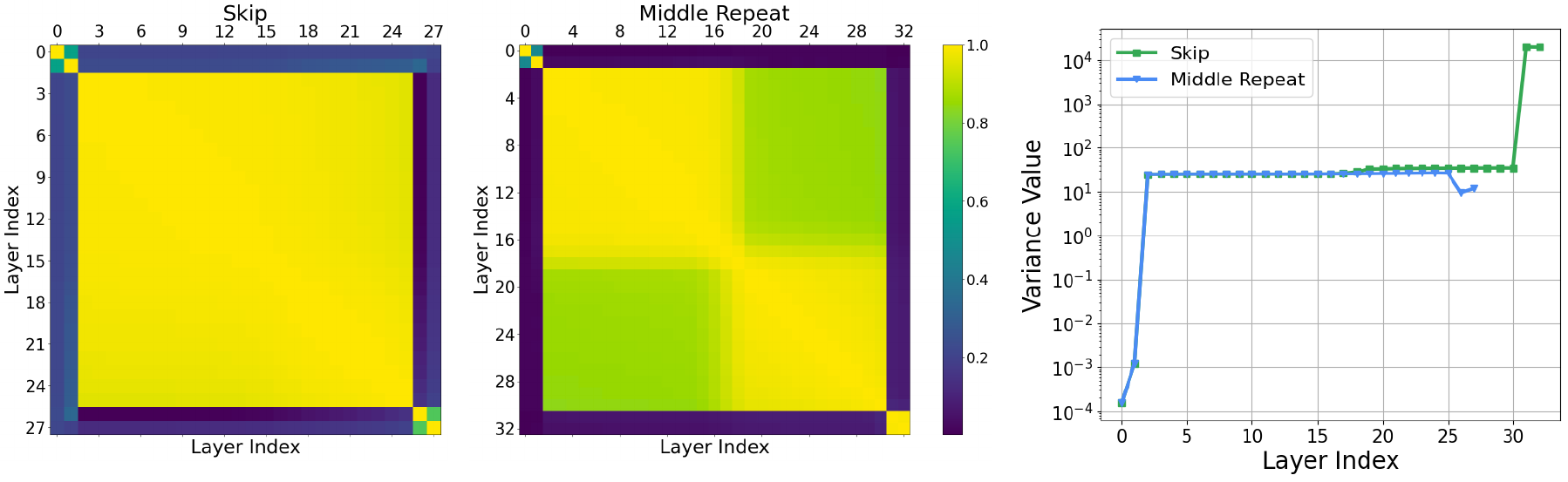}
  \caption{Comparing average cosine similarity and variance between the hidden states of all layers of Llama2-7B in Skip (top) and Middle Repeat (bottom)
    methods, the start layer is 13 therefore Skip uses 27 layers, skipping 5.}
  \label{figure:middle_repeat_vs_skip}
\end{figure*}

It's curious that replacing the middle layers with the weights of the center layer causes much worse performance than skipping these layers
altogether.  To help explain this, we analyzed the cosine similarity and statistical information for the hidden states of these two methods with
using start layer equal to 13 ($N=13$) for Llama2-7B.
In Figure \ref{figure:middle_repeat_vs_skip}, we find that the matrix for ``Skip'' shares the same cosine similarity trend with the full
Llama2-7B model, while in ``Middle Repeat'', the repeated center layer causes the hidden states to \emph{drifts} away from each other (the green
parts of the figure).  In addition, the ending layers of ``Middle Repeat'' behave differently from those for ``Skip'', their variance explore at last two layers.


\begin{figure*}[ht]
\begin{tabular}{rl} 
\begin{tabular}{c} 
  Baseline response
  \\ (Llama2-7B)
\end{tabular} 
&
\begin{tabular}{l} 
\scriptsize
\begin{boxedverbatim}
Mishka spent $16.50*3 = <<16.50*3=49.50>> 49.50 on shorts
Mishka spent $22.50*3 = <<22.50*3=67.50>> 67.50 on pants
Mishka spent $42*3 = <<42*3=126>> 126 on shoes
The total amount of money Mishka spent on all the clothing
  items is 49.50+67.50+126 = <<49.50+67.50+126=243>>243
#### 243
\end{boxedverbatim}
\normalsize
\end{tabular} 
\\
\\
\begin{tabular}{c} 
  Parallel Layers response
  \\ (Llama2-7B, N=14)
\end{tabular} 
&
\begin{tabular}{l} 
\scriptsize
\begin{boxedverbatim}
Mishka spent $16.50 * 3 = <<16.50*3=50.50>> 50.50 on the shorts.
$22.50 * 3 = <<22.50*3=70.50>> 70.50 on the pants.
$42 * 3 = <<42*3=142>> 142 on the shoes.
Mishka spent $50.50 + $70.50 + $142 =
  <<50.50+70.50+142>>192.50 on the clothing items.
#### 192.50
\end{boxedverbatim}
\normalsize
\end{tabular} 
\end{tabular} 
\caption{Responses to a question asking for the total cost of 3 sets of clothes from the full Llama2-7B model and the Parallel Layers version.
  Note that the setup is correct, but the errors are the arithmetical calculations.}
\label{figure:gsm8k}
\end{figure*}

From our painter analogy, this would be consistent with a painter additively drawing wheels several times on the same canvas, making the
painting dissimilar to what the painters above her have been trained on.  Or, to answer the question directly:

{\bf Because repeating a middle layer pushes the input out of the shared representation space.}

\subsection{Do the results hold for Mistral and Pythia?} 
\label{subsec:othermodels}

To further investigate the generalization of our findings, we include summaries our results for Mistral-7B \cite{jiang2023mistral7b} and
Pythia-6.9B.  Mistral-7B is a decoder-only model whose architecture is largely similar to Llama2-7B, but with some important modifications such
as a sliding window attention and grouped-query attention, and, of course, different weight values.  As the name suggests, Pythia-6.9B
\cite{biderman2023pythiasuiteanalyzinglarge} is a 6.9 billion parameter decoder model, trained on public data, with the same architecture as the
Open Pre-trained Transformer OPT-6.7B \cite{zhang2022optopenpretrainedtransformer}.

As can be seen in the figure below {\bf Mistral follows a remarkably similar trend as Llama2-7B}, which isn't surprising given the similarity of
the architectures.

\begin{figure}[h]
\includegraphics[width=\llamawidth]{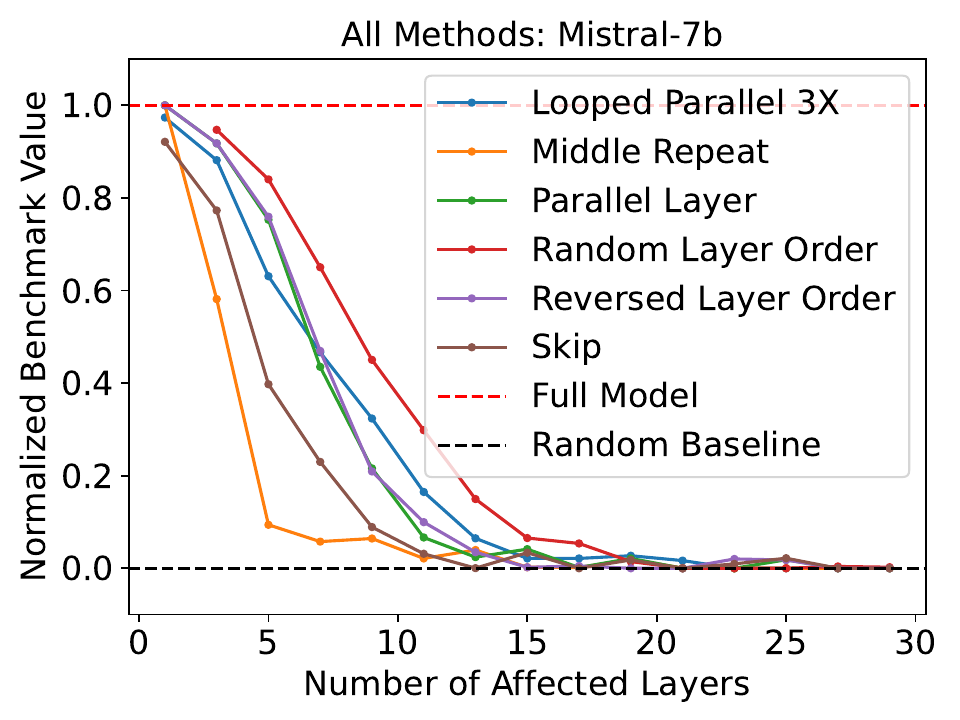}
\vspace{-2mm}
\end{figure}

However, in the figure below, we see that {\bf Pythia is less robust to modifications, especially modification of the Layer order} compared to
Mistral and Llama2-7B.

\includegraphics[width=\llamawidth]{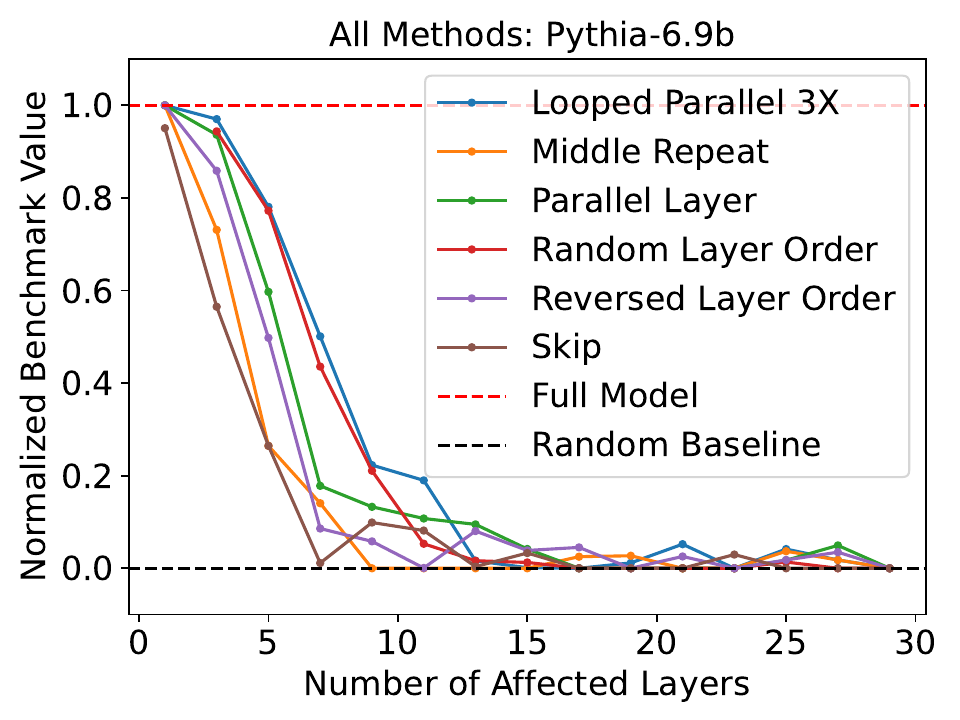}

\subsection{\bf Will internal looping improve over the base model?}

Can we shortcut the normal token-generation loop by doing this ``internally''?

\begin{figure}[ht]
  \centering
  \includegraphics[width=\llamawidth]{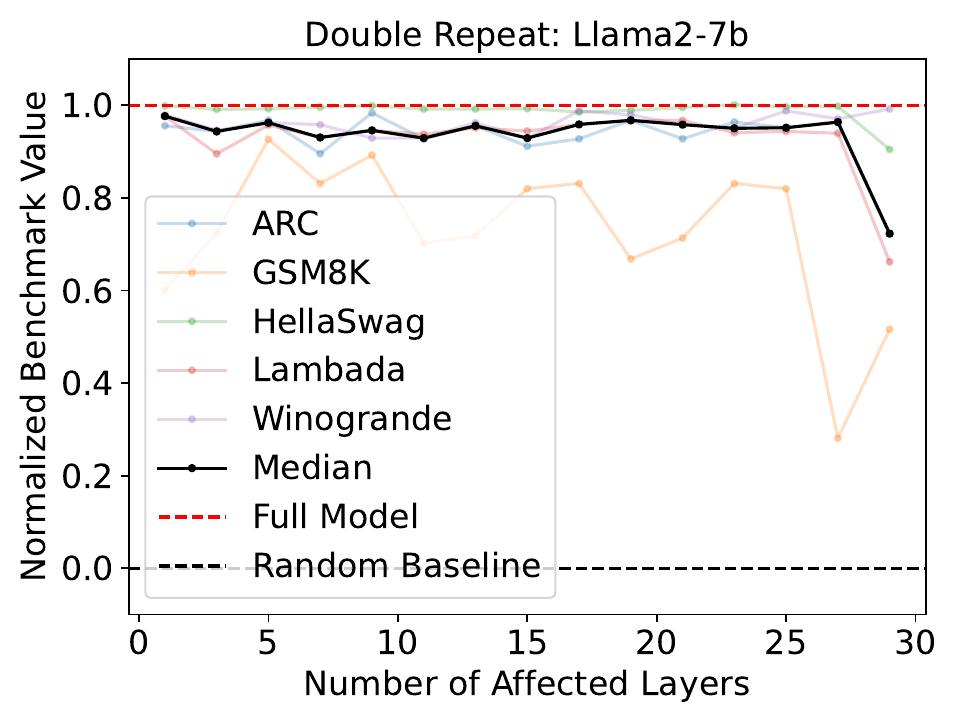}
  \includegraphics[width=\llamawidth]{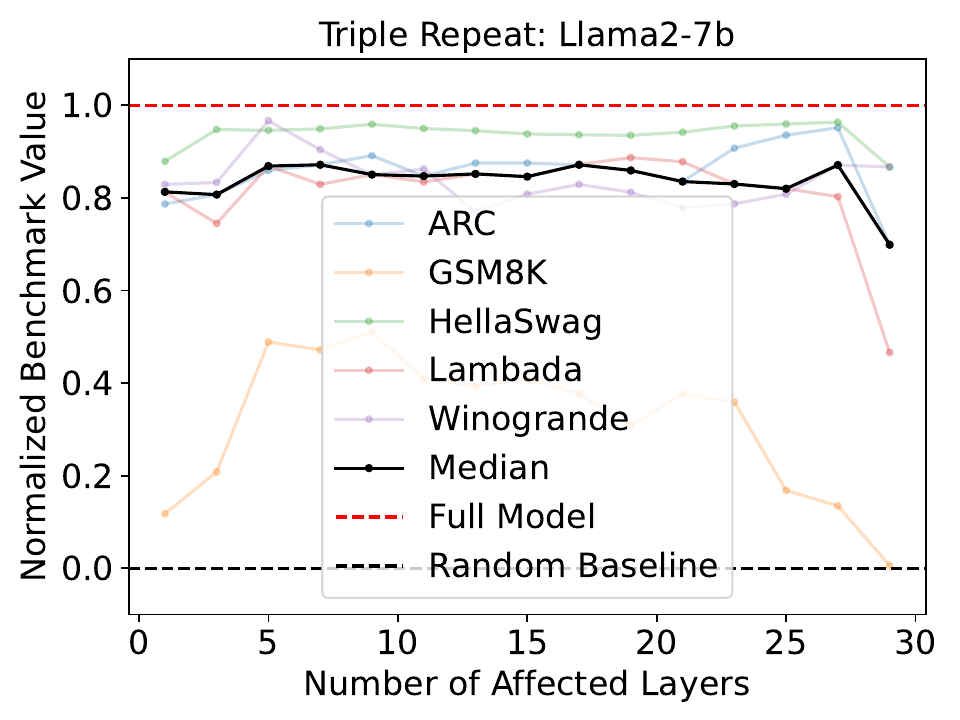}
  \caption{Repeating layers for Llama2-7B 2 and 3 times.}
  \label{figure:full_repeat}
\end{figure}

{\bf No. In Figure~\ref{figure:full_repeat}, we're still below the full model baseline.}

\subsection{Example of Error for Parallelized Layers}
\label{appendix:gsm8k}

One example from GSM8K might indicate that \emph{arithmetic} is especially dependent on layer order.  Figure \ref{figure:gsm8k} shows responses
to a question asking for the total cost of three sets of clothes.  The Parallel variant of Llama2-7B ($N=14$) sets up the correct calculations,
but errs in executing them correctly.

\subsection{BERT benchmarks}
\label{subsec:benchmarks}

Below are the benchmarks we used from GLUE \cite{wang-etal-2018-glue}:
\begin{description} 
\item[CoLA] (Corpus of Linguistic Acceptability): Acceptability judgments drawn from linguistic theory.
\item[MRPC] (Microsoft Research Paraphrase Corpus): Semantic equivalence for news sentences.
\item[QNLI] (Stanford Question Answering Dataset): Question answering from paragraphs.
\item[RTE] (The Recognizing Textual Entailment): Textual entailment
\item[SST2] (The Stanford Sentiment Treebank): Sentiment prediction.
\item[STSB] (The Semantic Textual Similarity Benchmark): Sentence pair similarity.
\item[WNLI] (The Winograd Schema Challenge): Sentence referent selection.
\end{description} 
Note that we did not include the QQP text classification nor the MNLI natural language inference benchmarks because of the high computational
cost of running these.

\subsection{Results for Frozen BERT}
\label{subsec:figures}
\vspace{-2mm}
Below are results for our BERT experiments using a fine-tuned head, but where the model parameters themselves are frozen.
Interestingly, both Frozen and Unfrozen Looped Parallel sometimes surpasses the full BERT model baseline.

\includegraphics[width=\llamawidth]{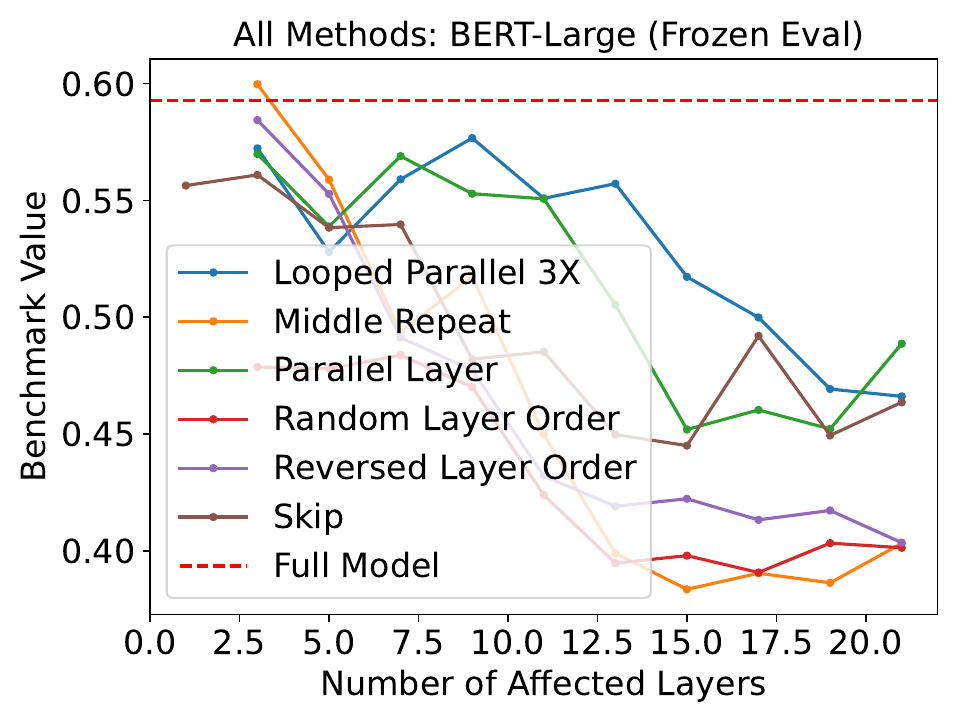}
\includegraphics[width=\llamawidth]{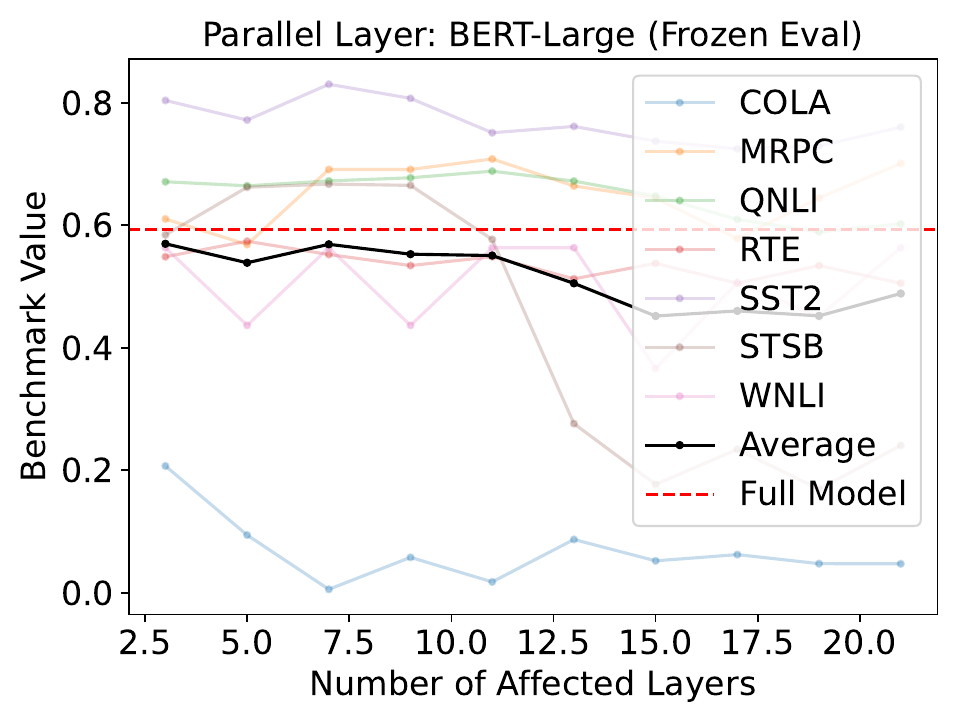}
\includegraphics[width=\llamawidth]{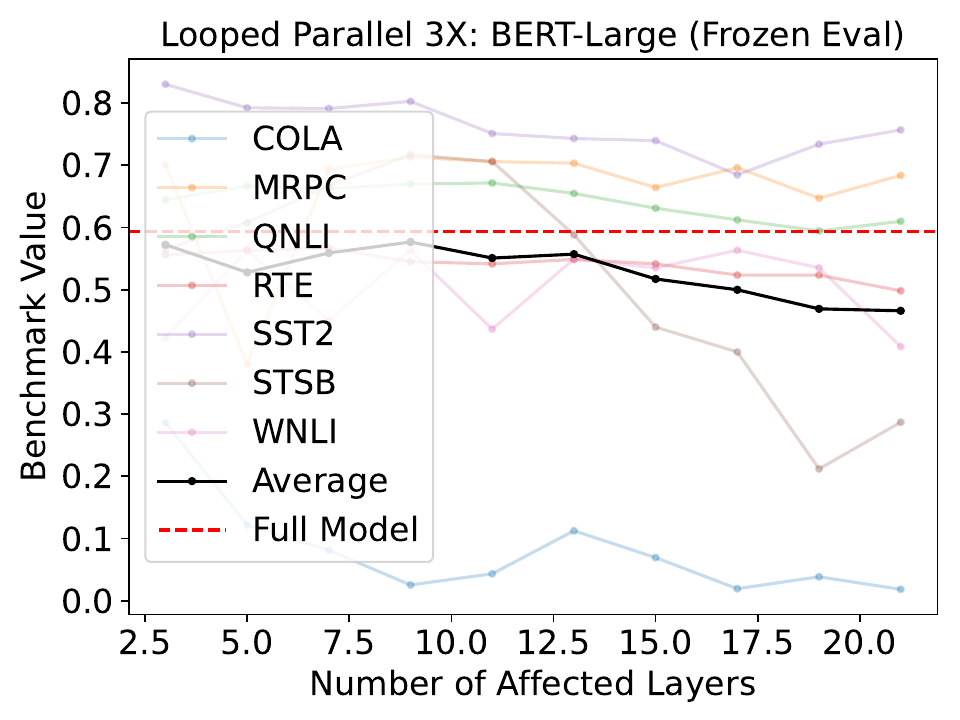}
\includegraphics[width=\llamawidth]{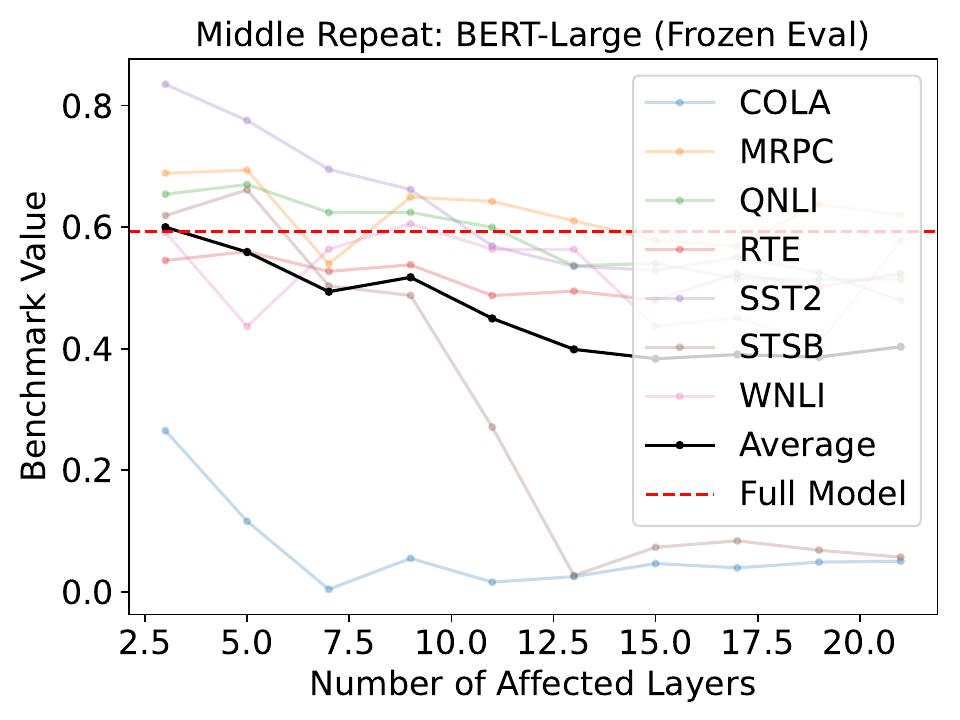}
\includegraphics[width=\llamawidth]{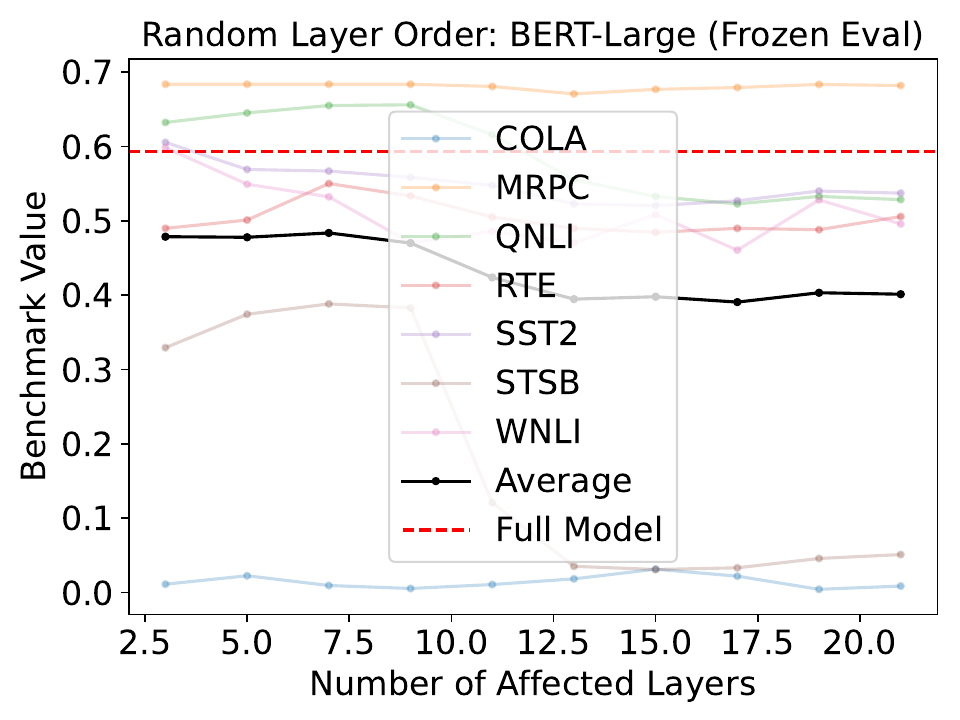}
\includegraphics[width=\llamawidth]{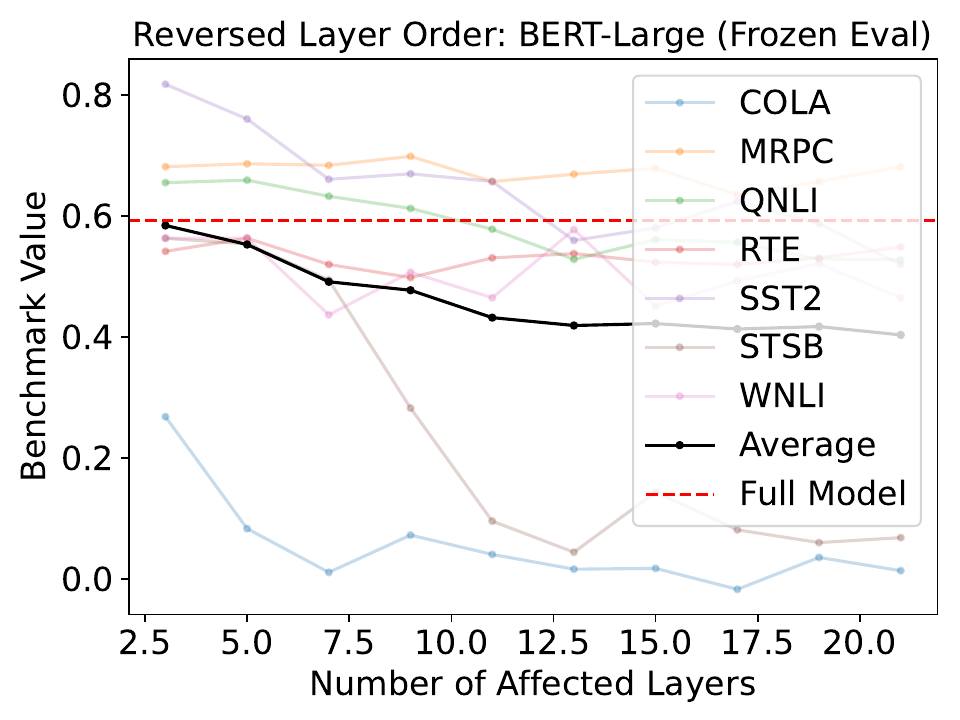}
\includegraphics[width=\llamawidth]{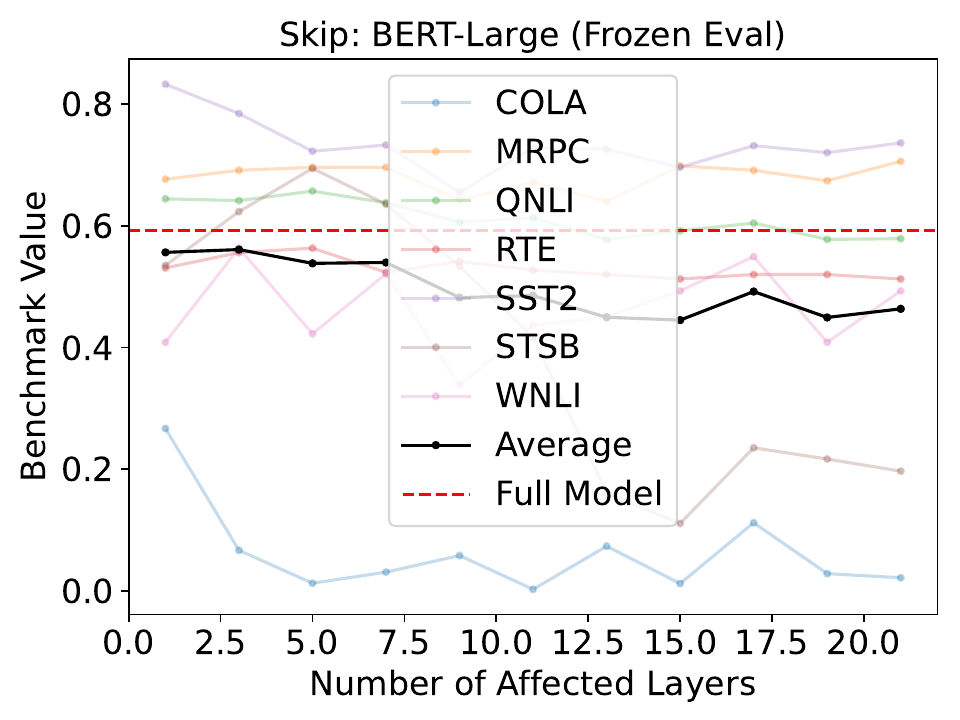}

\end{document}